\documentclass[runningheads]{llncs}

\usepackage{hyperref}
\usepackage{enumitem}
\usepackage{relsize}

\usepackage{mwe}
\usepackage[T1]{fontenc}
\usepackage{lmodern}
\usepackage{textcomp}

\usepackage{amsmath}
\usepackage{amsfonts}
\usepackage{mathtools}
\usepackage{scalerel,amssymb}
\usepackage{pifont}

\usepackage{graphicx}
\usepackage{subfigure}
\usepackage[usenames,dvipsnames]{color}
\usepackage{xcolor}
\usepackage{tcolorbox}

\usepackage{tikz}
\usetikzlibrary{arrows,arrows.meta,backgrounds,decorations.pathmorphing,positioning,fit,trees,shapes,shadows,automata,calc}
\pgfdeclarelayer{bg}    
\pgfsetlayers{bg,main}
\usepackage{pgfplots}
\usepackage{pgfplotstable}
\pgfplotsset{compat=newest}
\usepackage{dcolumn}
\usepackage{booktabs}

\usepackage{multirow}
\usepackage{makecell}
\usepackage{rotating}
\usepackage{multicol}
\usepackage{tablefootnote}
\usepackage{colortbl}
\usepackage{hhline}

\usepackage{url}
\usepackage{appendix}
\usepackage{footnote}
\usepackage{tablefootnote}
\usepackage{twoopt}
\usepackage{pdfpages}

\usepackage{caption}
\usepackage{lscape}
\usepackage{afterpage}

\usepackage{adjustbox}

\usepackage{xr}

\newcommand{\Paths}{\mbox{\sl Paths}}

\begin{document}
%
%
\title{Fine-Tuning the Odds in Bayesian Networks \thanks{This is a pre-copyedited version of a contribution published in Vejnarová J., Wilson N. (eds) Symbolic and Quantitative Approaches to Reasoning with Uncertainty. ECSQARU 2021. Lecture Notes in Computer Science, vol 12897. Springer, Cham, 268–283. The definitive authenticated version is available online via \textcolor{blue}{\url{https://doi.org/10.1007/978-3-030-86772-0_20}}. This work is funded by the ERC AdG Projekt FRAPPANT (Grant Nr. 787914).}
}

%
%
\author{Bahare Salmani\inst{1}\orcidID{0000-0003-3571-2502} \\
\and
Joost-Pieter Katoen\inst{1}\orcidID{0000-0002-6143-1926}}
\authorrunning{B. Salmani and J.-P. Katoen}
%
\institute{RWTH Aachen University, Aachen, Germany \\
\email{\{salmani,katoen\}@cs.rwth-aachen.de}}
\maketitle              
\begin{abstract}
This paper proposes new analysis techniques for Bayes networks in which conditional probability tables (CPTs) may contain symbolic variables.
The key idea is to exploit scalable and powerful techniques for synthesis problems in parametric Markov chains.
Our techniques are applicable to arbitrarily many, possibly dependent, parameters that may occur in multiple CPTs.
This lifts the severe restrictions on parameters, e.g., by restricting the number of parametrized CPTs to one or two, or by avoiding parameter dependencies between several CPTs, in existing works for parametric Bayes networks (pBNs). 
We describe how our techniques can be used for various pBN synthesis problems studied in the literature such as computing sensitivity functions (and values), simple and difference parameter tuning, ratio parameter tuning, and minimal change tuning.
Experiments on several benchmarks show that our prototypical tool built on top of the probabilistic model checker Storm can handle several hundreds of parameters.
\end{abstract}

\section{Introduction}

\paragraph{Parametric Bayesian networks.}
We consider Bayesian networks (BNs) whose conditional probability tables (CPTs) contain symbolic parameters such as $x_1$, $2{\cdot}x_1^2$, and $x_1{+}x_2$ with $0<x_1,x_2<1$.
Parametric probabilistic graphical models received a lot of attention, see e.g., \cite{coupe1998practicable,coupe2000sensitivity,druzdel2000building,DBLP:conf/ecsqaru/Jensen99,DBLP:journals/tsmc/Laskey95,DBLP:conf/ecsqaru/CastilloGH95,DBLP:conf/aaai/CastilloGH96,DBLP:journals/tsmc/CastilloGH97,DBLP:conf/uai/KjaerulffG00,DBLP:journals/jair/ChanD02,DBLP:conf/uai/ChanD04,DBLP:journals/amai/CoupeG02,DBLP:journals/ijar/Renooij14,DBLP:conf/ecsqaru/BoltG15}. 
Sensitivity analysis determines the effect of the parameter values in the CPTs on the decisions drawn from the parametric BN (pBN), e.g., whether $\Pr(H{=}h \mid E{=}e) > q$ for a given $q \in \mathbb{Q} \cap [0,1]$.
It amounts to establishing a function expressing an output probability in terms of the $x_i$ parameters under study. 
Parameter synthesis on pBNs deals with instantiating or altering the parameters such that the resulting BN satisfies some constraint of interest.
For pBNs, synthesis mostly amounts to parameter tuning: find the minimal change on the parameters such that some constraint, e.g., $\Pr(H{=}h \mid E{=}e) > q$ holds~\cite{DBLP:journals/ijar/ChanD05,DBLP:journals/tsmc/Laskey95}.
As sensitivity analysis and parameter synthesis are computationally hard in general \cite{DBLP:journals/tsmc/Laskey95,DBLP:conf/uai/KjaerulffG00,DBLP:conf/uai/KwisthoutG08},
many techniques restrict the number of parameters per CPT ($n$-way for small $n$~\cite{DBLP:journals/jair/ChanD02,DBLP:journals/amai/CoupeG02,van2007sensitivity}), permit parameter dependencies in several CPTs (single CPT~\cite{DBLP:conf/uai/ChanD04}), or consider specific structures such as join trees~\cite{DBLP:conf/uai/KjaerulffG00} and require all parameters to occur in the same clique of the junction tree. 

\paragraph{Parametric Markov chains.}
Quite independently, analysis techniques for Markov chains (MCs) have been developed in the formal verification community in the last two decades
\cite{DBLP:conf/ictac/Daws04,DBLP:journals/fac/LanotteMT07,DBLP:journals/sttt/HahnHZ11,DBLP:conf/cav/DehnertJJCVBKA15,DBLP:conf/atva/QuatmannD0JK16,DBLP:conf/atva/CubuktepeJJKT18,DBLP:conf/atva/GainerHS18,DBLP:conf/icse/FangCGA21}; for a recent overview see~\cite{DBLP:journals/corr/abs-1903-07993}.
Parametric MCs (pMCs) are MCs in which transitions are labelled with multivariate polynomials over a fixed set of parameters.
Substitution of these variables by concrete values induces a probability distribution over the state space of the MC.
Whereas early works focused on computing a rational function over the parameters expressing the reachability probability of a given target state, in the last years significant progress has been made to check whether there exists a parameter valuation inducing a MC that satisfies a given objective, or to partition the parameter space --- the space of all possible parameter values --- into ``good'' and ``bad'' w.r.t.\ a given objective, e.g., is the probability to reach some states below (or above) a given threshold $q$?
The complexity of various pMC synthesis problems is studied in~\cite{DBLP:journals/iandc/BaierHHJKK20,DBLP:journals/jcss/JungesK0W21}.

\begin{figure}[h]
\centering
  \begin{minipage}[b]{0.4\textwidth}
	\begin{tikzpicture}[
	node distance=0.5cm and 0.0cm,
	mNode/.style={draw,ellipse,align=center, minimum size=0.5cm}
	]
	\node[state,draw=none] (help){};
	
	\node[above=0.4cm of help] (tabUrine)
	{
		\begin{tabular}{c|rr}
        \multicolumn{1}{c}{} & \multicolumn{2}{c}{U} \\ 
        \cmidrule{2-3}
        \multicolumn{1}{c}{P} & \multicolumn{1}{c}{neg} & \multicolumn{1}{c}{pos} \\ 
        \hline\midrule
        no & 0.893 & 0.107 \\
        yes & p & 1-p \\
        \bottomrule
        \end{tabular}
	};
	
	\node[right=0.5cm of tabUrine] (tabBlood)
	{
		\begin{tabular}{c|rr}
        \multicolumn{1}{c}{} & \multicolumn{2}{c}{B} \\ 
        \cmidrule{2-3}
        \multicolumn{1}{c}{P} & \multicolumn{1}{c}{neg} & \multicolumn{1}{c}{pos} \\ 
        \hline\midrule
        no & 0.894 & 0.106 \\
        yes & q & 1-q \\
        \bottomrule
        \end{tabular}
	};
	\end{tikzpicture} 
	\end{minipage}
    $\qquad \qquad$
  \begin{minipage}{0.3\textwidth}
    \vspace*{-4.0cm}
    \includegraphics[scale=0.34]{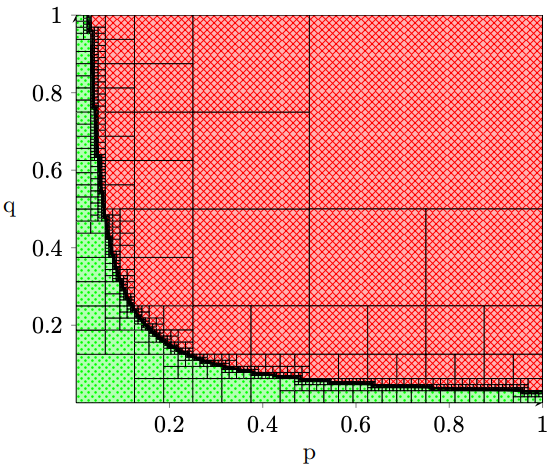}
  \end{minipage}
  \vspace*{-1.0cm}
  \caption{(left) Parametric CPTs and (right) the parameter space split into safe and unsafe regions for the constraint $\Pr(\text{P = yes} \mid \text{U = neg and B = neg}) \leq 0.2$.
  } 
  \label{motiv-red-green}
\end{figure}
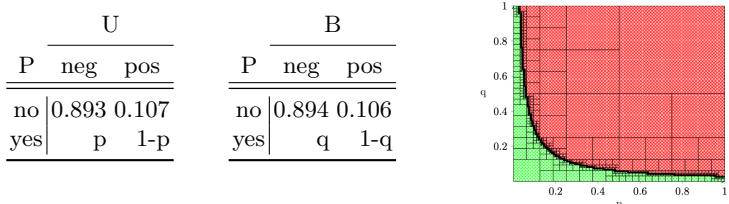

\emph{This paper aims to extend the spectrum of parameter synthesis techniques for parametric BNs, i.e., BNs in which arbitrary many CPTs may contain symbolic probabilities, with state-of-the-art and recently developed techniques for parametric MCs.} Consider the BN adapted from \cite{DBLP:books/daglib/0024906} depicted below.
\begin{figure}[htb!]
	\centering
    \resizebox{0.85\width }{0.85\height}{%
	\begin{tikzpicture}[
	node distance=1cm and 0.0cm,
	mNode/.style={draw,ellipse,align=center, minimum size=1cm}
	]
	\node[state,draw=none] (help){};
	\node[mNode,right=0.0cm of help] (v0) {Pregnancy};
	\node[mNode,below right=1.5cm of v0] (v2) {Blood Test};
	\node[mNode,below left=1.5cm of v0] (v3) {Urine Test};
	
	\node[left=0.5cm of v0] (tabPreg)
	{
		\begin{tabular}{rr}
        \multicolumn{2}{c}{Pregnancy} \\ 
        \midrule
        \multicolumn{1}{c}{no} & \multicolumn{1}{c}{yes} \\ 
        \hline\midrule
        0.13 & 0.87 \\
        \bottomrule
        \end{tabular}
	};
	\node[above left=-0.5cm and 0.5cm of v3] (tabUrine)
	{
		\begin{tabular}{c|rr}
        \multicolumn{1}{c}{} & \multicolumn{2}{c}{Urine Test} \\ 
        \cmidrule{2-3}
        \multicolumn{1}{c}{Pregnancy} & \multicolumn{1}{c}{neg} & \multicolumn{1}{c}{pos} \\ 
        \hline\midrule
        no & 0.893 & 0.107 \\
        yes & 0.36 & 0.64 \\
        \bottomrule
        \end{tabular}
	};
	
	\node[right=0.5cm of v0] (tabBlood)
	{
		\begin{tabular}{c|rr}
        \multicolumn{1}{c}{} & \multicolumn{2}{c}{Blood Test} \\ 
        \cmidrule{2-3}
        \multicolumn{1}{c}{Pregnancy} & \multicolumn{1}{c}{neg} & \multicolumn{1}{c}{pos} \\ 
        \hline\midrule
        no & 0.894 & 0.106 \\
        yes & 0.27 & 0.73 \\
        \bottomrule
        \end{tabular}
	};
	\path (v0) edge[-{Latex[length=2mm]}] (v3) 
	(v0) edge[-{Latex[length=2mm]}] (v2);	
	\end{tikzpicture}
	}
\end{figure}
\vspace*{-0.7cm}
The probability of a cow being pregnant given both tests are negative is about $0.45$. 
Assume the farmer wants to replace both tests such that this false-positive error is below $0.2$.
Fig.~\ref{motiv-red-green} (left) indicates the corresponding  pBN while (right) shows the synthesized values of the parameters $p$ and $q$ (the false-negative probabilities for the new urine and blood tests) using pMC techniques~\cite{DBLP:conf/atva/QuatmannD0JK16} such that the farmer's constraint is satisfied (green) or not (red). 

Let us highlight a few issues: we can treat parameter space partitionings that go beyond rectangular shapes, and we support multiple, possibly dependent, parameters (not illustrated in our example). 
Thanks to approximate techniques such as parameter lifting\cite{DBLP:conf/atva/QuatmannD0JK16}, the entire parameter space can be split into safe, unsafe, and --- by some approximation factor --- unknown regions.
This provides useful information: if it is not possible to find urine and blood tests of a certain quality, are alternatives fine too?
Parameter tuning~\cite{DBLP:journals/jair/ChanD02} for pBNs finds parameter values that are at a minimal distance to the original values in the BN.
For our example, the BN sensitivity tool SamIam suggests changing the false-negative probability of the urine test ($p$) from $0.36$ to $0.110456$; or changing the false-negative probability of the blood test ($q$) from $0.27$ to $0.082842$.\footnote{When it comes to \emph{multiple parameter suggestions}, SamIam suggests $p=0.120097$ or $q=0.089892$.} 
Interestingly, if the parameters occur in multiple CPTs, the constraint can be satisfied with a smaller deviation from the original parameter values.
Other recent work \cite{DBLP:conf/ictac/BartocciKS20} focuses on obtaining symbolic functions for pBN-related problems such as sensitivity analysis.  Note that pBNs are similar to constrained BNs~\cite{DBLP:journals/corr/BeaumontH17} that focus on logical semantics rather than synthesis algorithms and tools as we do. 

We treat the theoretical foundations of exploiting pMC techniques for various synthesis questions on pBNs, present a prototypical tool that is built on top of the probabilistic model checker Storm~\cite{DBLP:conf/cav/DehnertJK017} and the pMC analysis tool Prophesy~\cite{DBLP:conf/cav/DehnertJJCVBKA15}, and provide experimental results that reveal:
\begin{itemize}
\item pMC techniques are competitive to most common functions with the pBN tools SamIam and Bayesserver\footnote{We do not consider tools such as~\cite{DBLP:journals/aes/ToloPB18} for sensitivity analysis of Credal networks.}.
\item pMC techniques are well-applicable to general pBNs, in particular by allowing parameter dependencies.
\item Approximate parameter space partitioning is effective for parameter tuning e.g., ratio, difference, and minimal change problems.
\end{itemize}

Further proofs and details of this paper can be found in \cite{DBLP:journals/corr/abs-2105-14371}.


\section{Parametric Bayesian networks}
\label{sec-pbn}

This section defines \emph{parametric} Bayesian networks (BNs)\footnote{Our notion of parametric BN should not be confused with existing parametric notions that consider variations of the structure, i.e., the topology of the BN.} and defines the sensitivity analysis and parameter tuning tasks from the literature that we consider. 
\noindent
\emph{Parameters.}
Let $X = \{ x_1, \ldots, x_n \}$ be a finite set of real-valued \textit{parameters} and $\mathbb{Q}(X)$ denote the set of multivariate polynomials over $X$ with rational coefficients. 
A \emph{parameter instantiation} is a function $u \colon X \to \mathbb{R}$. A polynomial $f$ can be interpreted as a function $f \colon \mathbb{R}^n \to \mathbb{R}$ where $f(u)$ is obtained by substitution, i.e., in $f(u)$ each occurrence of $x_i$ in $f$ is replaced by $u(x_i)$.
To make clear where substitution occurs, we write $f[u]$ instead of $f(u)$ from now on.
We assume that all parameters are bounded, i.e., $\text{\it lb}_i \leq u(x_i) \leq \text{\it ub}_i$ for each parameter $x_i$.
The \emph{parameter space} of $X$ is the set of all possible parameter values, the hyper-rectangle spanned by the intervals $[\text{\it lb}_i, \text{\it ub}_i]$.
A subset $R$ of the parameter space is called a \emph{region}. \\[1ex]
\noindent
\emph{Variables.}
Let $V = \{v_1, \ldots, v_m\}$ be a set of variables. Let $D_{v_i}$ denote the domain of variable $v_i$. A \emph{joint variable valuation} is a function $\eta \colon V \to D_{v_1} \times \ldots D_{v_m}$ that maps each variable $v_i$ to some $d \in D_{v_i}$. Let $Eval(V)$ denote the set of all joint variable valuations for $V$. For the set $D$, let $Distr(D)$ denote the set of parametric probability distributions over $D$.  \\[1ex]
\noindent
\emph{Parametric Bayes networks.}
A parametric BN is a BN in which entries in the conditional probability tables (CPTs) are polynomials over the parameters in $X$.
\begin{definition}
\label{pBN-definition}
A \textit{parametric BN} (pBN) $\mathcal{B}$ is a tuple $(V,W,X,\Theta)$ with:
\begin{itemize}
\item $V=\{v_1,\ldots,v_m\}$ is a set of discrete \emph{random variables} with $\mbox{\sl dom}(v_i)=D_{v_i}$
\item $G=(V,W)$ with $W \subseteq V \times V$ is a \textit{directed acyclic graph} on $V$
\item $X = \{x_1, \dots, x_n\}$ is a finite set of real-valued \textit{parameters}
\item $\Theta$ is a set of \textit{parametric conditional probability tables} $\Theta=\{\, \Theta_{v_i} \mid v_i\in V \, \}$.
Let $parents(v_i)$ denote the set of parents for the node $v_i$ in $G$. The CPT for variable $v_i$ is the function $\Theta_{v_i} \colon Eval(parents(v_i)) \to \mbox{\it Distr}(D_{v_i})$ that maps each evaluation $\overline{par} \in Eval(parents(v_i))$ to a parametric probability distribution $\Theta_{v_i}(\overline{par})$ over $D_{v_i}$. The \emph{CPT entry} $\Theta_{v_i}(\overline{par})(d_i)$ denotes the probability of $v_i = d_i$ given the parents evaluation $\overline{par}$.
\end{itemize}
\end{definition}
Let $\mathcal{B}[u]$ be obtained by replacing every parameter $x_i$ in $\mathcal{B}$ by its value $u(x_i)$.
A parameter instantiation $u$ is \emph{well-formed} for the pBN $\mathcal{B}$ if $\mathcal{B}[u]$ is a BN, i.e., for every $v_i \in V$ and parent evaluation $\overline{par}$, $\Theta_{v_i}(\overline{par})[u]$ yields a probability distribution over $D_{v_i}$. In the sequel, we assume $u$ to be well-formed. A pBN defines a \emph{parametric} joint probability distribution over $V$.\\[1ex]
\noindent
\emph{pBN subclasses.} 
We define some sub-classes of pBNs that are used in existing sensitivity analysis techniques and tools. 
They constrain the number of parameters, the number of CPTs (and the number of rows in a CPT) containing parameters. 
Let $\mathcal{B}=(V,W,X,\Theta)$ be a pBN, $c(x_i)$ the number of CPTs in $\mathcal{B}$ in which $x_i$ occurs and $r(x_i)$ the number of CPT rows in which $x_i$ occurs.
\begin{itemize}
\item 
$\mathcal{B} \in p_1c_1r_1$ iff $\mathcal{B}$ contains one parameter $x_1$ and $x_1$ only occurs in a single row of a single CPT, i.e., $X = \{ x_1 \}$, $c(x_1) = r(x_1) = 1$. 
\item 
$\mathcal{B} \in p_2c_{\leq 2}r_1$ iff $\mathcal{B}$ involves two parameters occurring only in two rows of two (or one) CPTs, i.e., $X = \{ x_1, x_2 \}$, $c(x_i) \in \{ 1,2 \}$, $r(x_i) = 1$ for $i=1,2$. 
\item $\mathcal{B} \in p_*c_1r_1$ iff $\mathcal{B}$ allows multiple distinct parameters, provided each parameter occurs in a single row of a single CPT, i.e., $r(x_i) = 1$, $c(x_i) = 1$ for each $x_i$ and all the parameters occur in the same CPT.
\end{itemize}
The class $p_1c_1r_1$ is used in one-way, $p_2c_{\leq 2}r_1$ in two-way sensitivity analysis~\cite{DBLP:journals/jair/ChanD02,DBLP:journals/amai/CoupeG02,van2007sensitivity} and $p_*c_1r_1$in single CPT~\cite{DBLP:conf/uai/ChanD04}.
\\[1ex]
\noindent
\emph{Parameter synthesis problems in pBN.}
\label{pbn-problems}
We define some synthesis problems for pBNs by their corresponding decision problems~\cite{DBLP:conf/uai/KwisthoutG08}.
Let $\Pr$ denote the parametric joint distribution function induced by pBN $\mathcal{B}=(V,W,X,\Theta)$ and $\Pr[u]$ the joint probability distribution of $\mathcal{B}[u]$ at well-formed instantiation $u$. 
Let $E \subseteq V$ be the evidence, $H \subseteq V$ the hypothesis and $q \in \mathbb{Q} \, \cap \, [0,1]$ a threshold.
\begin{description}
\item[Parameter tuning.] Find an instantiation $u$ s.t.
$\Pr[u](H=h \mid E=e) \ \geq \ q.$
\item[Hypothesis ratio parameter tuning.] Find an instantiation $u$ s.t.
\[ 
\frac{\Pr[u](H=h' \mid E=e)}{\Pr[u](H=h \mid E=e)} \ \geq \ q 
\quad \mbox{i.e.,} \quad
\frac{\Pr[u](H=h', E=e)}{\Pr[u](H=h, E=e)} \ \geq \ q,
\]
where $h$ and $h'$ are joint variable evaluations for the hypothesis $H$.
\item[Hypothesis difference parameter tuning.] Find an instantiation $u$ s.t.
\[
\Pr[u](H=h \mid E=e) - \Pr[u](H=h' \mid E=e) \ \geq \ q,
\]
where $h$ and $h'$ are joint variable evaluations for the hypothesis $H$.
\item[Minimal change parameter tuning.]
For a given parameter instantiation $u_0$ and $\epsilon \in \mathbb{Q}_{> 0}$, find an instantiation $u$ s.t. 
\[
d(\Pr[u], \Pr[u_0]) \ \leq \ \epsilon,
\]
where $d$ is a distance notion on probability distributions, see~\cite{DBLP:journals/ijar/ChanD05}.
\item[Computing sensitivity function and sensitivity value.]
For the evidence $E=e$ and the hypothesis $H=h$, compute the sensitivity function:
    \begin{align*}
        f_{\Pr(H=h\,|\,E=e)} \ = \ \Pr(H=h \mid E=e).
    \end{align*}
This is a rational function over $X$, i.e., a fraction $g/h$ with $g,h \in \mathbb{Q}(X)$.
\end{description}
The difference and ratio problems can analogously be defined for evidences. 
The evidence tuning problems are defined for values $e$ and $e'$ for $E$, given a fixed value $h$ for $H$.

\section{Parametric Markov chains}
\label{sec-pmc}

A parametric Markov chain is a Markov chain in which the transitions are labelled with polynomials over the set $X$ of parameters. These polynomials are intended to describe a parametric probability distribution over the pMC states.
\begin{definition}
\label{paramMC}
A \textit{parametric Markov chain} (pMC) $\mathcal{M}$ is a tuple $(S,s_0,X,P)$ where $S$ is a finite set of \textit{states} with \textit{initial state} $s_0\in S$, $X$ is as before, and
$P: S \times S\to\mathbb{Q}(X)$ is the transition probability function.
\end{definition}
For pMC $\mathcal{M}=(S,s_0,X,P)$ and well-formed parameter instantiation $u$ on $X$, $\mathcal{M}[u]$ is the discrete-time Markov chain $(S,s_0,X,P[u])$ where $P[u]$ is a probability distribution over $S$. We only consider well-formed parameter instantiations. \\[1ex]
\noindent
\emph{Reachability probabilities.}
Let $\mathcal{D}$ be an MC.
Let $\Paths(s)$ denote the set of all infinite paths in $\mathcal{D}$ starting from $s$, i.e., infinite sequences of the form $s_0 s_1 s_2 \ldots$ with $P(s_i,s_{i+1}) > 0$.
A probability measure $\Pr_{\mathcal{D}}$ is defined on measurable sets of infinite paths using a standard cylinder construction; for details, see, e.g.,~\cite[Ch.~10]{BK08}.
For $G \subseteq S$, let
\begin{equation}\label{eqn:probmeasure}
\Pr_{\mathcal{D}}(\lozenge G) \ = \ \Pr_{\mathcal{D}} \{ \,s_0 s_1 s_2 \ldots \in \Paths(s_0) \mid \exists i. \, s_i \in G  \, \}
\end{equation}
denote the probability to eventually reach some state in $G$ from $s_0$.
For pMC $\mathcal{M}$, $\Pr_{\mathcal{M}}(\lozenge G)$ is a function with $\Pr_{\mathcal{M}}(\lozenge G)[u] = \Pr_{\mathcal{D}}(\lozenge G)$ where $\mathcal{D} = \mathcal{M}[u]$, see~\cite{DBLP:conf/ictac/Daws04}.
\\[1ex]
\noindent
\emph{Parameter synthesis problems on pMCs.}
We consider the following synthesis problems on pMCs.
Let $\mathcal{M}=(S,s_0,X,P)$ be a pMC and $\lambda \in \mathbb{Q} \cap [0,1]$ a threshold, $\sim$ a binary comparison operator, e.g.,  $<$ or $\geq$, and $G \subseteq S$.
\begin{description}
\item[Feasibility problem.] Find a parameter instantiation $u$ s.t.\ 
$\Pr_{\mathcal{M}[u]}(\lozenge G) \ \sim \ \lambda$.
\item[Synthesis problem.] Partition a region $R$ into $R_a$ and $R_r$ s.t.\
\[
\displaystyle \Pr_{\mathcal{M}[u]}(\lozenge G) \ \sim \ \lambda \mbox{ for all } u \in R_a 
\quad \mbox{and} \quad
\displaystyle  \Pr_{\mathcal{M}[u]}(\lozenge G) \ \not\sim \ \lambda \mbox{ for all } u \in R_r.
\]
$R_a$ is called an accepting region and $R_r$ a rejecting region. 
\item[Approximate synthesis problem.]
Partition a region $R$ into an accepting region $R_a$, rejecting region $R_r$, and unknown region $R_u$, such that $R_a \cup R_r$ covers at least $c\%$ of $R$. Additionally, $R_a$,$R_r$, and $R_u$ should be finite unions of rectangular regions. 
\item[Verification problem.] Check whether region $R$ is accepting, rejecting, or inconsistent, i.e., neither accepting nor rejecting.
\item[Computing reachability functions.]
Compute the rational function $\Pr_{\mathcal{M}}(\lozenge G)$.
\end{description}

\paragraph{Algorithms for pMC synthesis problems.}
Several approaches have been developed to \emph{compute the reachability function} $\Pr_{\mathcal{M}}(\lozenge G)$. 
This includes state elimination~\cite{DBLP:conf/ictac/Daws04}, fraction-free Gaussian elimination~\cite{DBLP:journals/iandc/BaierHHJKK20} and decomposition~\cite{DBLP:conf/qest/JansenCVWAKB14,DBLP:journals/corr/abs-2102-01490}.
The reachability function can grow exponentially in the number of parameters, even for acyclic pMCs~\cite{DBLP:journals/iandc/BaierHHJKK20}.
\emph{Feasibility} is a computationally hard problem: finding parameter values for a pMC that satisfy a reachability objective is ETR-complete\footnote{Existential Theory of the Reals. ETR problems are between NP and PSPACE, and ETR-hard problems are as hard as finding the roots of a multi-variate polynomial.}~\cite{DBLP:journals/jcss/JungesK0W21}\!.
Feasibility has been tackled using sampling search methods such as PSO\footnote{Particle swarm optimization.}
 and Markov Chain Monte Carlo~\cite{DBLP:conf/tase/ChenHHKQ013} and solving a non-linear optimization problem~\cite{DBLP:conf/tacas/BartocciGKRS11}. 
State-of-the-art approaches~\cite{DBLP:conf/tacas/Cubuktepe0JKPPT17,DBLP:conf/atva/CubuktepeJJKT18} iteratively simplify the NLP\footnote{Nonlinear programming.} 
 encoding around a point to guide the search.  
The \emph{approximate synthesis problem} checking is best tackled with parameter lifting~\cite{DBLP:conf/atva/QuatmannD0JK16}. 
The parameter lifting algorithm (PLA) first drops all dependencies between parameters in a pMC.
It then transforms the pMC into a Markov decision process to get upper and lower bounds for the given objective.

\externaldocument{introduction}
\newcommand{\B}{\mathcal{B}}
\newcommand{\M}{\mathcal{M}}
\newcommand{\D}{\mathcal{D}}
\newcommand{\Hbefore}{H_{\mbox{\it \scriptsize before}}}
\newcommand{\Hafter}{H_{\mbox{\it \scriptsize after}}}
\newcommand{\Hvars}{H_{\mbox{\it \scriptsize vars}}}
\newcommand{\Evars}{E_{\mbox{\it \scriptsize vars}}}
\newcommand{\Elast}{E_{\mbox{\it \scriptsize last}}}

\section{Analysing parametric BNs using pMC techniques}
\label{sec-transformation}

The key of our approach to tackle various synthesis problems on pBNs is to exploit pMC techniques.
To that end, we transform a pBN into a pMC.
We first present a recipe that is applicable to all inference queries on pBNs, and then detail a transformation that is tailored to the evidence in an inference query. \\[1ex]
\noindent
\emph{A pBN2pMC transformation.}
This is inspired by our mapping of BNs onto tree-shaped MCs \cite{DBLP:conf/qest/SalmaniK20}. 
Here, we propose an alternative transformation that yields more succinct (p)MCs, as it only keeps track of a subset of (p)BN variables at each ``level'' of the (p)MC.
Intuitively, the value of variable $v_i$ is still remembered if it has already been processed and at least one of its children has not. 

\paragraph{Notations. }Let $\varrho = (v_1, \ldots, v_m)$ be a topological order over the variables in $V$ with respect to the DAG $G$ and $\varrho(v)$ denote the index of variable $v$ in the order. Let $E = E_1 \wedge \ldots E_k$ be the evidence, where $E_i$ is in the form of $v_{E_i} = d_{E_i}$ with $v_{E_i} \in V$ and
\begin{align*}
v_{E_1} <_{\varrho} v_{E_2} <_{\varrho} \ldots <_{\varrho} v_{E_k} \quad \text{ i.e., }\quad \varrho(v_{E_1}) < \varrho(v_{E_2}) < \ldots < \varrho(v_{E_k}).
\end{align*}
In the sequel, we often denote $E_k$ as $E_{\mbox{\it \scriptsize last}}$.

\begin{figure}
\centering
  \begin{minipage}[b]{0.65\textwidth}
      \resizebox{0.6\width }{0.6\height}{%
        \begin{tikzpicture}[
		node distance=0.3cm and 0.3cm,
		mNode/.style={draw,ellipse,align=center, minimum size=0.5cm},
		mLNode/.style={align=center, minimum size=0.5cm}
		]
		\node[](dummy){};
		\node[mNode,below=0.5cm of dummy] (init){init};
		\node[mNode,below left=1.cm and 0.3cm of init] (t1) {P=no};
		\node[mNode,below right=1.cm and 0.3cm of init] (f1) {P=yes};
		
		\node[mNode,below=1cm of t1] (t1t3) {P=no \\ U=pos};
		\node[mNode,left=1cm of t1t3] (t1f3) {P=no \\ U=neg};
		\node[mNode,below=1cm of f1] (f1t3) {P=yes \\ U=neg};
		\node[mNode,right=1cm of f1t3] (f1f3) {P=yes \\ U=pos};

		\node[mNode,below right=2cm and 0.5cm of t1f3] (t2) {B=neg};
		\node[mNode,below left=2cm and 0.5cm of f1f3] (f2) {B=pos};
		
		\draw [-{Latex[length=2mm]}] (dummy) -- (init) node[midway, above, sloped] {};
		\draw [-{Latex[length=2mm]}] (init) -- (t1) node[midway, above, sloped] {$0.13$};
		\draw [-{Latex[length=2mm]}] (init) -- (f1) node[midway, above, sloped] {$0.87$};
		
		\draw [-{Latex[length=2mm]}] (t1) -- (t1t3) node[midway, above, sloped] {$0.107$};
		\draw [-{Latex[length=2mm]}] (t1) -- (t1f3) node[midway, above, sloped] {$0.893$};
		\draw [-{Latex[length=2mm]}] (f1) -- (f1t3) node[midway, above, sloped] {$p$};
		\draw [-{Latex[length=2mm]}] (f1) -- (f1f3) node[midway, above, sloped] {$1{-}p$};	
		
		\draw [-{Latex[length=2mm]}] (t1t3) -- (t2) node[pos=0.65, above, sloped] {$0.894$};
		\draw [-{Latex[length=2mm]}] (t1t3) -- (f2) node[pos=0.15, above, sloped] {$0.106$};
		\draw [-{Latex[length=2mm]}] (t1f3) -- (t2) node[pos=0.35, above, sloped] {$0.894$};
		\draw [-{Latex[length=2mm]}] (t1f3) -- (f2) node[pos=0.1, above, sloped] {$0.106$};	
		\draw [-{Latex[length=2mm]}] (f1t3) -- (t2) node[pos=0.7, above, sloped] {$q$};
		\draw [-{Latex[length=2mm]}] (f1t3) -- (f2) node[pos=0.5, above, sloped] {$1{-}q$};
		\draw [-{Latex[length=2mm]}] (f1f3) -- (t2) node[pos=0.15, above, sloped] {$q$};
		\draw [-{Latex[length=2mm]}] (f1f3) -- (f2) node[pos=0.55, above, sloped] {$1{-}q$};
		
		\path[every loop/.append style=-{Latex[length=2mm]}]
		(t2) edge [loop below] node {1} (t2)
		(f2) edge [loop below] node {1} (f2);
	\end{tikzpicture}
	}
    \end{minipage}
    \hspace{-2cm}
  \begin{minipage}[b]{0.3\textwidth}
      \resizebox{0.6\width }{0.6\height}{%
        \begin{tikzpicture}[
		node distance=0.3cm and 0.3cm,
		mNode/.style={draw,ellipse,align=center, minimum size=0.5cm},
		mLNode/.style={align=center, minimum size=0.5cm}
		]
		\node[](dummy){};
		\node[mNode,below=0.5cm of dummy] (init){init};
		\node[mNode,below left=1.cm and 0.3cm of init] (t1) {P=no};
		\node[mNode,below right=1.cm and 0.3cm of init] (f1) {P=yes};
		
		\node[mNode,below=1cm of t1] (t1t3) {P=no \\ U=neg};
		\node[mNode,below=1cm of f1] (f1t3) {P=yes \\ U=neg};
		\node[mNode,below=1cm  of t1t3] (t2) {P=no \\ B=neg};
		\node[mNode,below=1cm of f1t3] (f2) {P=yes \\ B=neg};
		
			\draw [-{Latex[length=2mm]}] (dummy) -- (init) node[midway, above, sloped] {};
		\draw [-{Latex[length=2mm]}] (init) -- (t1) node[midway, above, sloped] {$0.13$};
		\draw [-{Latex[length=2mm]}] (init) -- (f1) node[midway, above, sloped] {$0.87$};
		
		\draw [-{Latex[length=2mm]}] (t1) -- (t1t3) node[midway, above, sloped] {$0.893$};
		
		 \draw [-{Latex[length=2mm]},dotted] (t1) to [bend left=45] node [above, sloped] (TextNode1) {$0.107$} (init);

		\draw [-{Latex[length=2mm]}] (f1) -- (f1t3) node[midway, above, sloped] {$p$};
		
		 \draw [-{Latex[length=2mm]},dotted] (f1) to [bend right=45] node [above, sloped] (TextNode1) {$1{-}p$} (init);

		\draw [-{Latex[length=2mm]}] (t1t3) -- (t2) node[midway, above, sloped] {$0.894$};
		
		 \draw [-{Latex[length=2mm]},dotted] (t1t3) to [bend left=95] node [above, sloped] (TextNode1) {$0.106$} (init);			
		
		\draw [-{Latex[length=2mm]}] (f1t3) -- (f2) node[midway, above, sloped] {$q$};

		 \draw [-{Latex[length=2mm]},dotted] (f1t3) to [bend right=95] node [above, sloped] (TextNode1) {$1{-}q$} (init);
		
		\path[every loop/.append style=-{Latex[length=2mm]}]
		(t2) edge [loop below] node {1} (t2);
				\path[every loop/.append style=-{Latex[length=2mm]}]
		(f2) edge [loop below] node {1} (f2);
	\end{tikzpicture}
	}
    \end{minipage}
    \caption{The generated pMCs for the pregnancy test example based on (left) pBN2pMC and (right) evidence-guided pBN2pMC transformation.}
    \label{fig:transformation-example}
\end{figure}
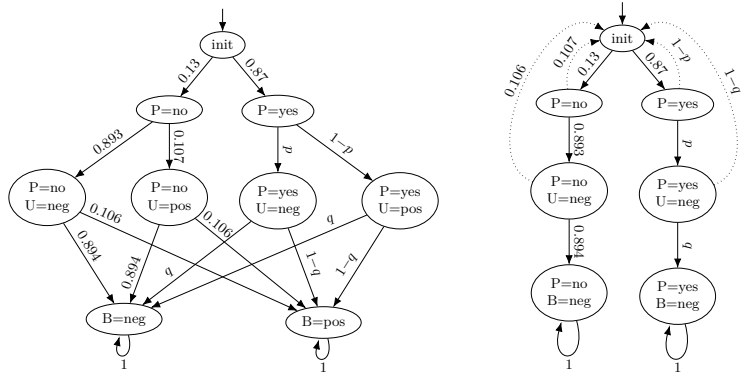

\noindent The hypothesis $H$ can be in the form any arbitrary logical formula over the atomic propositions $v_i = d _i$.
W.l.o.g, let $H = H_1 \wedge \ldots \wedge H_l$ be the hypothesis, where $H_i$ is in the form of $v_{H_i} = d_{H_i}$ and $\varrho(v_{H_i}) < \varrho(v_{H_{i+1}})$. For the logical formula $A$, let $\mbox{\it vars}(A)$ denote the set of variables that appear in $A$.
Let $*$ denote the \emph{don't care} value. 
For the variable $v_i \in V$, let $D^*_{v_i} = D_{v_i} \cup \{*\}$. A \emph{variables state} is a function $s \colon D^*_{v_1} \times \ldots \times D^*_{v_m}$ that maps each variable $v_i \in V$ either to some $d_i$ or to the \emph{don't care} value. Intuitively speaking, $v_i = *$ if the value of variable $v \in V$ is either not yet determined or not needed any more. Let $\mbox{\it States}(V)$ denote the set of all variables states. Let $s \in \mbox{\it States}(V)$.
\begin{itemize}
\item We write $s \models (v_i = d_i)$ iff $s(v_i) = d_i$ with $d_i \in D_{v_i}$.
\item We have $s \models (v_i = s(v_i) \wedge \ldots \wedge v_m = s(v_m))$.
\item For $V' \subseteq V$ and $s' \in Eval(V')$, we write
\begin{align*}
s \models s' \text{ iff } \bigwedge_{v \in V'} s(v_i) = s'(v_i).
\end{align*}
\item For the logical formula $\alpha$ over the atomic propositions $(v_i = d_i)$, we write
\begin{align*}
s \models \alpha \text{ iff }  (v_i = s(v_i) \wedge \ldots \wedge v_m = s(v_m)) \implies \alpha.
\end{align*}
\end{itemize}
Given the pBN $\mathcal{B} = (V,W,X,\Theta)$, the topological order $\varrho$ on $V$, and the state $s \in \mbox{\it States}(V)$, let $s[v_i{=}d_i] = s^*$, where
\begin{itemize}
\item $s^* \in \mbox{\it States}(V)$,
\item $s^*(v_i) = d_i$,
 \item $s^*(v) = s(v)$ for $v \neq v_i$ satisfying
 \begin{align*}
 \varrho(v) < \varrho(v_i) \text{ and } \exists c \in children(v) \,.\, \varrho(c) > \varrho(v_i), \text{ and}
 \end{align*}
 \item $s^{*}(v) = *$ for all other variables. 
\end{itemize}
\begin{definition}[Evidence-agnostic pMC of pBN]
\label{defRMC}
Let $\mathcal{B}=(V,W,X, \Theta)$ be a pBN and $\varrho = (v_1, \dots, v_m)$ be a topological order over $V$. The \textit{pMC} of $\mathcal{B}$ is $\mathcal{M}^{\varrho}_{\mathcal{B}}=(S, s_0, X, P)$, where:
\begin{itemize}
\item $s_0 =(v_1=*, \ldots, v_m=*)$ is the initial state,
\item $S \subseteq \mbox{\it States}(V)$ is the set of states defined as follows:
\begin{itemize}
\item $S = \bigcup\limits_{0 \leq i \leq m} S_i$
\item $S_0 = \{s_0\}$
\item $S_{i{+}1} = \{ s_{i}[v_{i{+}1}{=}d_{i{+}1}] \,\mid\, \text{for } s_{i} \in S_{i} \text{ and } d_{i{+}1} \in D_{v_{i{+}1}}\} \text{ for } 0 \leq i < m$
\end{itemize}
\item  and $P \colon S \times S \to [0,1]$ is the transition probability functions defined by the following two rules.
\end{itemize}
\begin{enumerate}
\item Let $0 \leq i < m$, $\overline{par} \in \mbox{\it Eval}(\mbox{\it parents}(v_{i{+}1}))$, and $s_{i{+}1} = s_{i}[v_{i{+}1}{=}d_{i{+}1}] \in S_{i{+}1}$. For each $s_{i} \in S_{i}$ and $d_{i{+}1} \in D_{v_{i{+}1}}$,
\begin{align*}
P(s_{i}, s_{i{+}1}) = \Theta(\overline{par})(d_{i{+}1}) \text{ iff } s_{i} \models \overline{par}.
\end{align*}
\item $P(s_m, s_m) = 1$ for each $s_m \in S_m$.
\item $P(s_i, s_j) = 0$, otherwise.
\end{enumerate}
\end{definition}
The states generation in Definition \ref{defRMC} ensures that 
\begin{align*}
\forall 0 \leq i < m, \forall s_{i} \in S_{i}\,\cdot\, \exists! \,\overline{par} \in \mbox{\it Eval}(\mbox{\it parents}(v_{i{+}1})) \text{ such that } s_{i} \models \overline{par}.
\end{align*}
\noindent
\paragraph{Example.} Fig. \ref{fig:transformation-example} (left) indicates the pMC obtained by Def. \ref{defRMC} for the pregnancy test pBN and the topological ordering $(P, U, B)$. The node names are abbreviated by their first letters and the ``don't care'' evaluations are omitted.  The pMC starts in the initial state $s_0$, where all the variables are ``don't care''. The successor states of $s_0$ are determined by the first node in the topological ordering, i.e., \emph{Pregnancy} ($P$): for each value in the domain of $P$, $s_0$ has a successor state. This yields the states $P=yes$ and $P=no$ in the pMC. The transition probabilities from $s_0$ to these states naturally come from the the CPT of \emph{Pregnancy}, see rule 1. The states in the next \emph{level}\footnote{Note that Def. \ref{defRMC} imposes no backward edges: (a) $P(s_i, s_j) > 0$ iff $s_i \in S_i$ and $s_j \in S_{i{+}1}$ or $i=j=m$, (b) $S_i \cap S_j = \emptyset$ for $i \neq j$. This allows us to consider \emph{levels} for $\M^{\varrho}_{\B}$: we refer to the states in $S_i$ as the states in the $i$'th level of the pMC.} are subsequently determined by the node \emph{Urine Test}. The transition probability, e.g., from the state $P{=}no$ to the state $(P{=}no,U{=}neg)$ is determined by the corresponding CPT entry that is $0.893$. Note that the value of pregnancy is still kept at this level as $P$ still has an unprocessed child ($B$). In the last level, the value of Pregnancy is forgotten as there is no children of $P$ remaining in the topological order.
\noindent
\paragraph{}The following result relates (conditional) inference in pBNs to (conditional) reachability objectives in pMCs. The proof is provided in App.~\ref{app:proofs}.
\begin{proposition} 
\label{eq:query-correspondence}
Let $\B$ be a pBN and $\varrho$ be a topological ordering on $V$. Let $E$ and $H$ be the evidence and the hypothesis. Then, for the pMC $\M^{\varrho}_{\B}$ obtained from Def. \ref{defRMC} we have:
\[
\Pr_{\B}(E) \ = \ 1 - \Pr_{\M^{\varrho}_{\B}}(\lozenge \, \neg E) 
\quad \text{and} \quad
\Pr_{\B} (H \mid E) \ = \ \dfrac{1 - \Pr_{\M^\varrho_\B}(\lozenge\,(\neg H \lor \neg E))}{1 - \Pr_{\M^{\varrho}_{\B}}(\lozenge \, \neg E)},
\]
where the latter equality requires $\Pr_{\B}(\neg E) < 1$.
\end{proposition}
This result directly enables to use techniques for \emph{feasibility checking} on pMCs to pBNs, and the use of techniques to compute reachability functions on pMCs to \emph{computing sensitivity functions} on pBNs. \\[1ex]
\noindent
\emph{An evidence-tailored pBN2pMC transformation.}
The above transformation is agnostic from the inference query.
We now construct a pMC $\M^\varrho_{\B, E} $ tailored to a given evidence $E$.
The transformation is inspired by a transformation on MCs~\cite{DBLP:conf/tacas/BaierKKM14} for computing conditional reachability probabilities.
Let $\mathcal{B} = (V,E,X,\Theta)$ be a pBN and $\varrho$ be a topological order on $V$. 
Let $E = (v_{E_1} = d_{E_1}) \wedge \cdots \wedge (v_{E_k}= d_{E_k})$ be the evidence defined as before.
We construct pMC $\M^\varrho_{\B, E}$ by the following two amendments on the pMC $\M^\varrho_{\B}$ as defined in Def.~\ref{defRMC}:
\begin{itemize}
\item \emph{Propagation operation: }For $v_j \not \in \mbox{\it vars}(E)$ with $\varrho(v_j) < \varrho(v_{E_k})$, we propagate the values of $v_j$ until the level $\varrho(v_{E_k})$. \\
\item \emph{Redirection operation: } Let $\M^{\downarrow}_{\B, \varrho} = (S^{\downarrow}, s^{\downarrow}_0, X, P^{\downarrow})$ be the pMC obtained from applying the propagation operation on $\M^{\varrho}$. Let $S_{\neg E} = \{s^{\downarrow} \in S^{\downarrow} \, |\, s^{\downarrow} \models (v_{E_i} = \neg d_{E_i}) \text{ for some } 1 \leq i \leq k \text{ and } \neg d_{E_i} \neq d_{E_i}  \}$ be the set of sates in $\M^{\downarrow}_{\B, \varrho}$ that violate the evidence $E$. We reroute the direct transitions to the states in $S_{\neg E}$ to the initial state $s_0$ and delete the states in $S_{\neg E}$.
\end{itemize}
\noindent
Let $H$ be a logical formula over the atomic propositions $(v_{H_i}=d_{H_i})$ that is decomposable to the sub-formulas $\Hbefore$ and $\Hafter$, such that the variables appearing in $\Hbefore$ all occur before $v_{\Elast}$ in the topological order $\varrho$ and the variables involved in $\Hafter$ all occur after $v_{\Elast}$. Without loss of generality, let ${H} = (v_{H_1}{=}d_{H_1}) \wedge \cdots \wedge (v_{H_l}{=}d_{H_l})$ be the hypothesis. Let $H_{\mbox{\it \scriptsize before}} = \bigwedge_{i=1}^{b} (v_{H_i}{=}d_{H_i})$ with $v_{H_i} <_{\varrho} v_{E_{\mbox{\it \scriptsize last}}}$ and let $H_{\mbox{\it \scriptsize after}} = \bigwedge_{i=a}^{l} (v_{H_i}{=} d_{H_i})$ with $v_{H_i} >_{\varrho} v_{E_{\mbox{\it \scriptsize last}}}$, such that $H = \Hbefore \wedge \Hafter$.
\begin{proposition}
\label{proposition-ev-tailored}
For the evidence-tailored pMC $\M^{\varrho}_{\B, E}$ of pBN $\B$, we have:
\[
          \Pr_{\B}(H \mid E) \ =  \  1 - \Pr_{\M^{\varrho}_{\B, E}}\big(\lozenge \big( \, (\neg \Hbefore \wedge \Elast) \lor \neg \Hafter \big)\big).
\]
\end{proposition}

\noindent
This result facilitates using pMC techniques for pBN \emph{parameter tuning}. For the formal definitions and the proof, see App.~\ref{app:proofs}.
\noindent
\paragraph{Example.} Fig. \ref{fig:transformation-example} (right) indicates the evidence-guided pMC generated for our running example, the ordering $(P, U, B)$, and the evidence $\sl U{=}neg \wedge B{=}neg$. By the propagation operation the value of $P$ is propagated until the level of last evidence node $B$. By the redirection operation, the transitions leading to $U{=}pos$ and $B{=}pos$ are redirected to the initial state. By Proposition \ref{proposition-ev-tailored},
\begin{align*}
\Pr_{\B}(P{=}yes\,|\, U{=}neg \wedge B{=}neg) = 1 - \Pr_{\M_{\B, E}}(\lozenge (P{=}no \wedge B{=}neg)).
\end{align*}
\noindent
\emph{Ratio and difference parameter tuning by parameter lifting.}
The ratio problem on pBN $\B$ corresponds to finding an instantiation $u$ in the pMC $\M^\varrho_{\B, E}$ s.t.
\begin{equation}
\label{ratio-constraint}
\Pr_{\M^{\varrho}_{\B, E}}[u](\lozenge T) \ \geq \
q \cdot \Pr_{\M^{\varrho}_{\B, E}}[u](\lozenge G),
\end{equation}
where $\Pr(\lozenge T)$ stands for $1 - \Pr_{\M^\varrho_{\B, E}}[u](\lozenge (H = \neg h' \lor E = \neg e))$ and $\Pr(\lozenge G)$ abbreviates $1 - \Pr_{\M^\varrho_{\B, E}}[u](\lozenge (H = \neg h \lor E = \neg e))$.
The problem can be solved using PLA: let region $R \subseteq \mathbb{R}_{\geq 0}^n$. 
We perform PLA for reaching $G$ and reaching $T$ on $\mathcal{M}_{\mathcal{B}}$, respectively.
This gives upper ($UB_T$, $UB_G$) and lower bounds ($LB_T$, $LB_G$) for the probabilities on the left- and the right-hand side of (\ref{ratio-constraint}).
Then:
\begin{itemize}
    \item If $LB_T \geq q \cdot UB_G$, the region $R$ is accepting for the ratio property.
    \item If $LB_T \leq q \cdot UB_G$, the region $R$ is rejecting for the ratio property.
    \item Otherwise, refine the region $R$.
\end{itemize}
For difference parameter tuning, we adopt the above recipe by replacing  (\ref{ratio-constraint}) by:
\begin{equation}
\label{diff-constraint}
\Pr_{\M^\varrho_{\B, E}}[u](\lozenge T) \ \geq \
q + \Pr_{\M^\varrho_{\B, E}}[u](\lozenge G).
\end{equation}

\section{Experiments}
\label{sec-experiments}
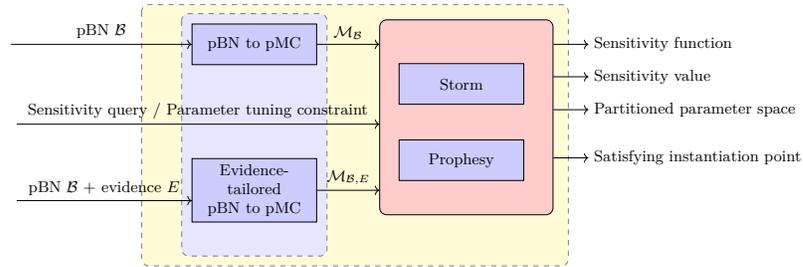
\begin{figure}[t!]
\begin{adjustbox}{width=0.95\columnwidth,center}
\newcommand{\mx}[1]{\mathbf{\bm{#1}}} 
\newcommand{\vc}[1]{\mathbf{\bm{#1}}} 

\pagestyle{empty}

\pgfdeclarelayer{background}
\pgfdeclarelayer{foreground}
\pgfsetlayers{background,main,foreground}

\tikzstyle{sensor}=[draw, fill=blue!20, text width=7em, 
    text centered, minimum height=2.5em]
\tikzstyle{ann} = [above, text width=5em]
\tikzstyle{naveqs} = [sensor, text width=10em, fill=red!20, 
    minimum height=12em, rounded corners]
\def\blockdist{2.5}
\def\edgedist{3.0}
\def\edgeshortdist{0.7}

\begin{tikzpicture}
\centering

    \node (dummy){dummy};
    \node (naveq)[naveqs] {}; 
    \node (st) [above right=0.0cm and -2.0 of dummy][sensor] {Storm};
   \node (proph) [below=0.7cm of st][sensor] {Prophesy};

    \path (naveq.140)+(-\blockdist,0) node (pBN) [sensor] {pBN to pMC};
    \path (naveq.-140)+(-\blockdist,0) node (accel) [sensor] {Evidence-tailored \\ pBN to pMC};
    
    \node (dummy2)[below left=2.6cm and 3.5 of pBN]{};

    \path [draw, <-] (pBN) -- node [above] {
    pBN $\mathcal{B}$} 
        (dummy2|- pBN.east);
    \path [draw, ->] (dummy2) -- node [above]{pBN $\mathcal{B}$ + evidence $E$}
        (accel.south west |- dummy2);        
    
        

    \node (dummy-mid)[above=1.3cm of dummy2]{};

    \path [draw, ->] (dummy-mid) -- node [above]{Sensitivity query / Parameter tuning constraint}
        (naveq.west |- dummy-mid);

    \path [draw, ->] (pBN) -- node [above] {$\mathcal{M}_{\mathcal{B}}$} 
        (naveq.west |- pBN) ;
    \path [draw, ->] (accel) -- node [above] {$\mathcal{M}_{\mathcal{B}, E}$} 
        (naveq.west |- accel);
    \node (Tools) [below of=accel] {};
    \path (naveq.south west)+(-0.6,-0.4) node (INS) {};
    \draw [->] (naveq.40) -- node [ann] {} + (\edgeshortdist,0) 
        node[right] {Sensitivity function};
    \draw [->] (naveq.25) -- node [ann] {} + (\edgeshortdist,0) 
        node[right] {Sensitivity value};
    \draw [->] (naveq.5) -- node [ann] {} + (\edgeshortdist,0)
        node [right] {Partitioned parameter space};
    \draw [->] (naveq.-25) -- node [ann] {} + (\edgeshortdist,0) 
        node[right] {Satisfying instantiation point};
    
    \begin{pgfonlayer}{background}
        \path (pBN.west |- naveq.north)+(-1.0,0.3) node (a) {};
        \path (INS.south -| naveq.east)+(+0.3,-0.5) node (b) {};
        \path[fill=yellow!20,rounded corners, draw=black!50, dashed]
            (a) rectangle (b);
        \path (pBN.north west)+(-0.2,0.2) node (a) {};
        \path (Tools.south -| pBN.east)+(+0.2,-0.2) node (b) {};
        \path[fill=blue!10,rounded corners, draw=black!50, dashed]
            (a) rectangle (b);
    \end{pgfonlayer}
\end{tikzpicture}
\end{adjustbox}
 \caption{Our prototypical tool-chain for synthesis problems on pBNs}
 \label{fig:tool-chain}
\end{figure}

\begin{figure}[h]
\centering
 \includegraphics[scale=0.65]{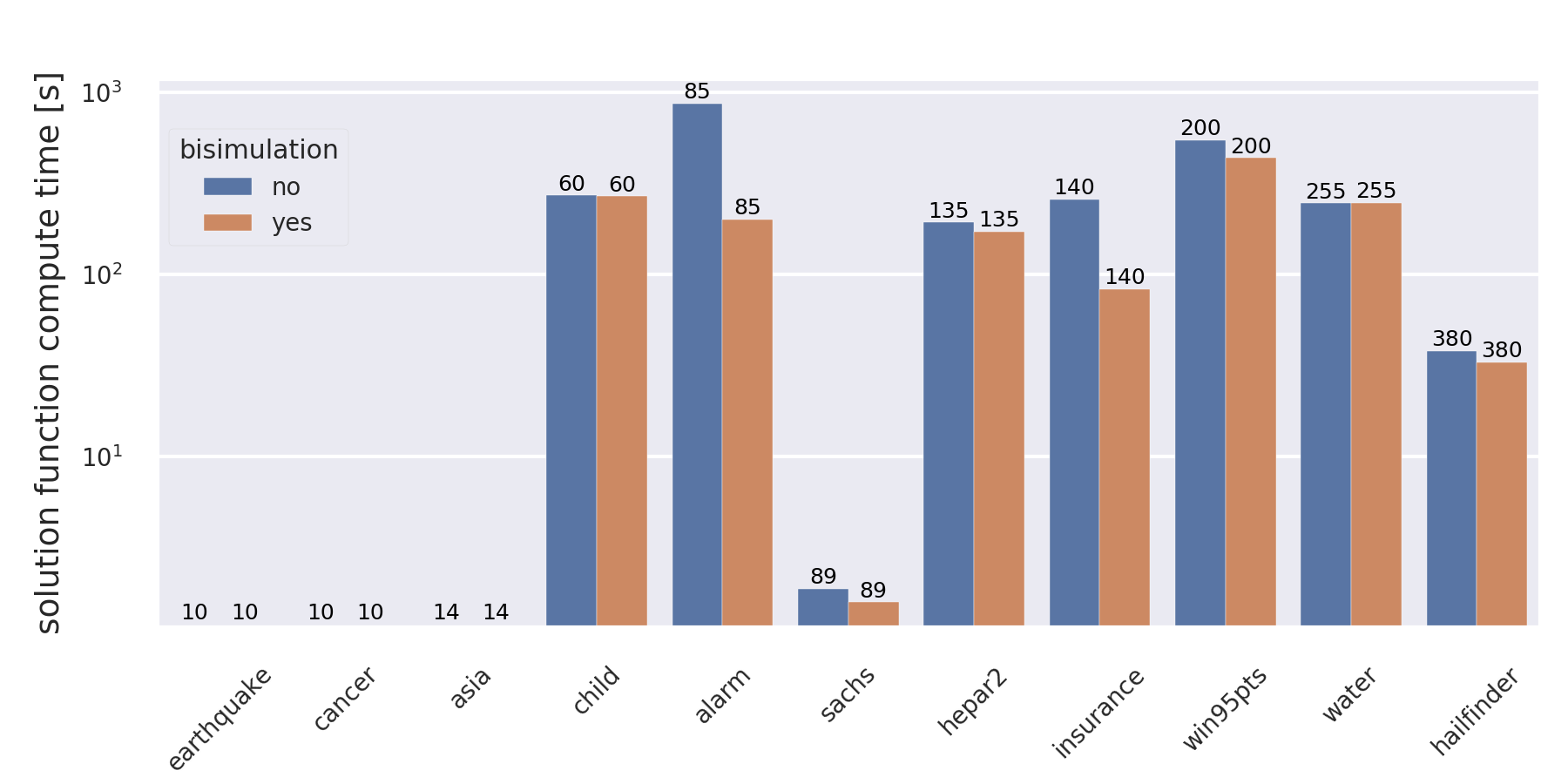}
 \caption[Storm's performance for calculating prior sensitivity functions of pBNs.]{Storm's performance for calculating prior sensitivity functions of pBNs.\footnotemark}
 \label{solution-function-chart}
\end{figure}

\begin{figure}[t!]
\hspace{-5mm}
  \begin{minipage}[b]{0.45\textwidth}
    \includegraphics[scale=0.23]{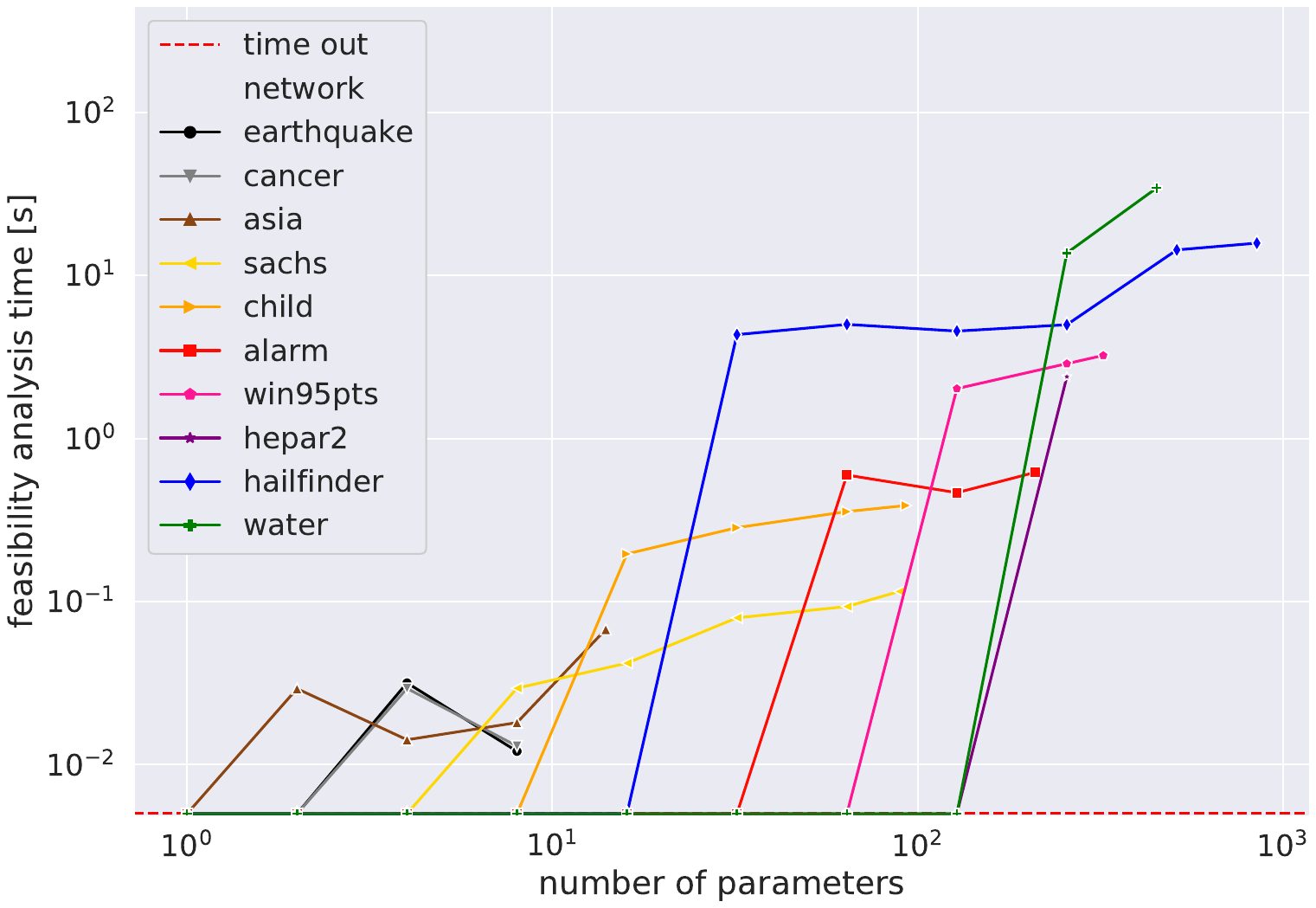}
  \end{minipage}
  \hspace{9mm}
  \begin{minipage}[b]{0.45\textwidth}
    \includegraphics[scale=0.23]{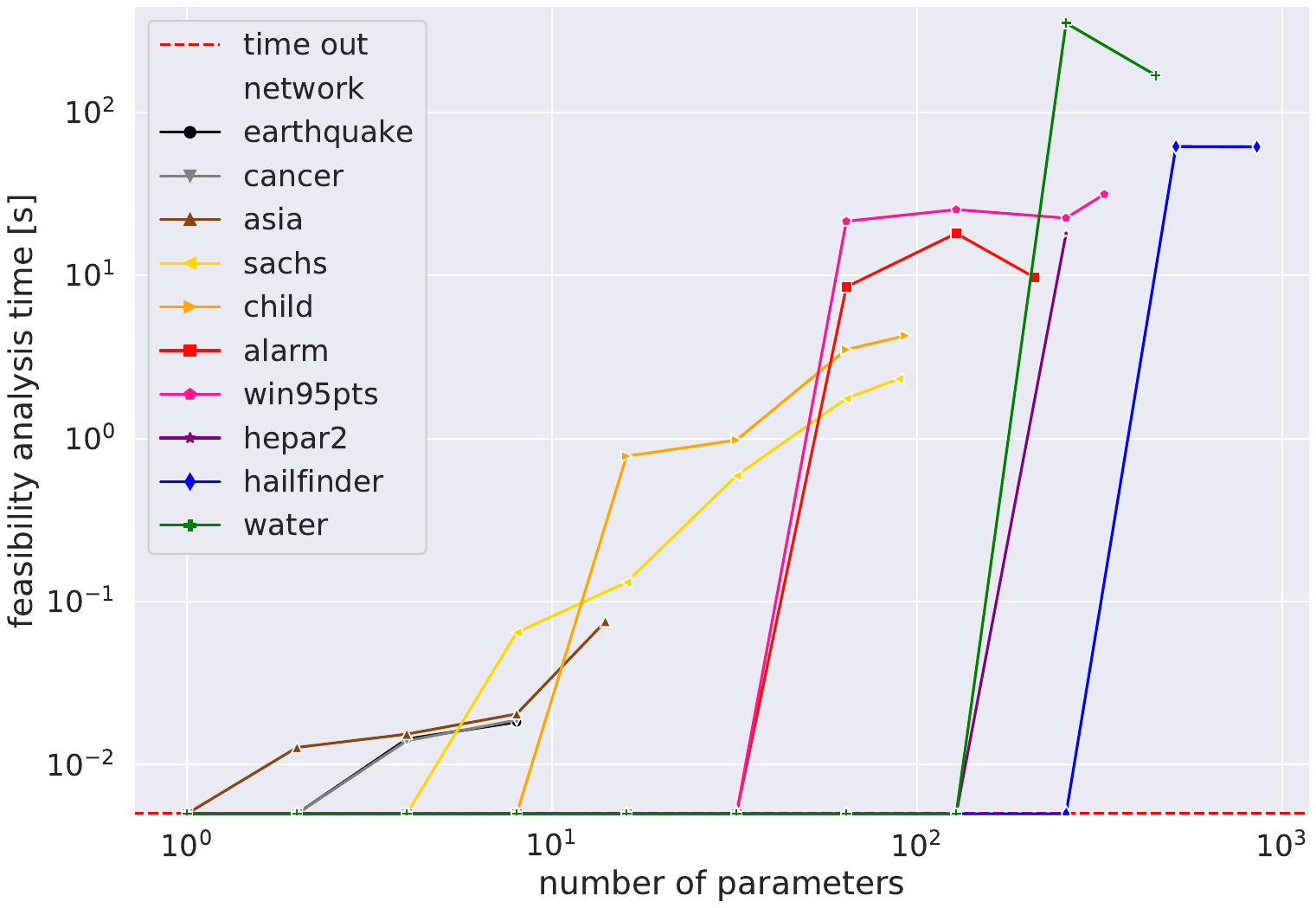}
  \end{minipage}
  \caption{Feasibility checking on pBN benchmarks by (left) QCQP and (right) PSO.} \label{feasibility-charts}
  \end{figure}

\begin{figure}[h]
  \centering
  \begin{minipage}[b]{0.45\textwidth}
    \includegraphics[scale=0.32]{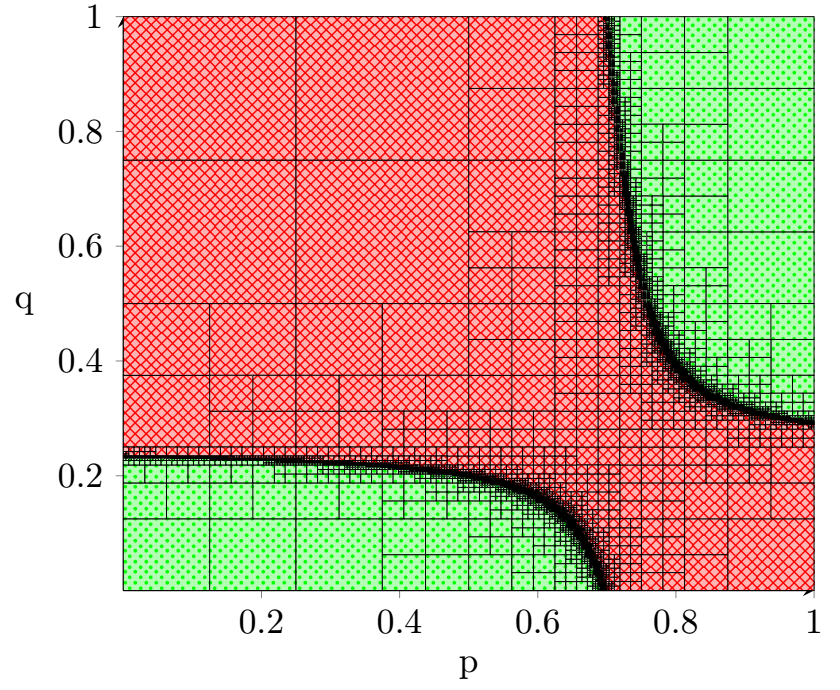}
  \end{minipage}
  \hspace{9mm}
  \begin{minipage}[b]{0.45\textwidth}
    \includegraphics[scale=0.34]{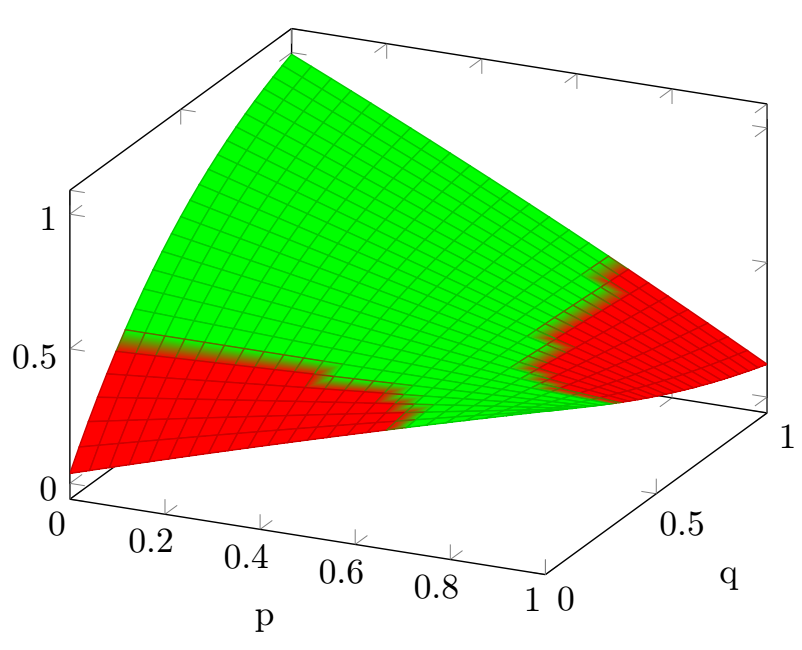}
  \end{minipage}
 \caption{PLA results on the \texttt{alarm} pBN ($p_2c_3r_{26}$) for the constraint $\Pr(venttube=0 \mid ventlung=0) > 0.6$ with a $99\%$ parameter space coverage.}  \label{alarm-pla-2d}
  \end{figure}

\emph{Our pBN analysis tool.}
We developed a prototypical tool on top of the tools Storm \cite{DBLP:conf/cav/DehnertJK017} and  Prophesy \cite{DBLP:conf/cav/DehnertJJCVBKA15}, see Fig.~\ref{fig:tool-chain}.
Storm is a probabilistic model checker that dominated the last (and only) two model-checking competitions, see \url{qcomp.org}; Prophesy is an efficient tool for pMC synthesis.
Our tool deploys pMC parameter synthesis techniques to analyze pBNs. 
It includes both pBN2pMC transformations where pBNs are provided in an extended \texttt{bif} format. 
The pMCs are either encoded in Jani \cite{DBLP:conf/tacas/BuddeDHHJT17} or in the explicit \texttt{drn} format. 
It is also possible to transform non-parametric BNs into MCs and parameterize the MC. 
Storm is used to compute the sensitivity function and for parameter tuning using PLA. 
Prophesy is exploited for feasibility checking: find a parameter instance satisfying an inference query.
Our tool-chain supports $p_*c_*r_*$, the general pBNs class.
As baseline we used two synthesis tools for parametric BNs: SamIam and Bayesserver. 
\\[1ex]
\noindent
\emph{SamIam.}
SamIam\footnote{\url{http://reasoning.cs.ucla.edu/samiam}} is a commonly used tool for the sensitivity analysis for pBNs, developed at Darwiche's group at UCLA. 
It allows the specification of conditional, hypothesis ratio, and hypothesis difference constraints on pBNs.
SamIam then attempts to identify minimal parameter changes that are necessary to satisfy these constraints. 
It supports the pBN classes $p_1c_1r_1$ and $p_*c_1r_1$. \\[1ex]
\emph{Bayesserver.} 
Bayeserver\footnote{\url{https://www.bayesserver.com}} is a commercial tool that offers sensitivity analysis and parameter tuning of pBNs.
For sensitivity analysis, it computes the sensitivity function and sensitivity value. 
It also performs minimal-change parameter tuning for conditional, hypothesis ratio, and hypothesis difference constraints. 
It supports the classes $p_1c_1r_1$ and $p_2c_{\leq 2}r_1$ for sensitivity analysis and the class $p_1c_1r_1$ for parameter tuning. Tab.~\ref{tab:tools-features} lists the functionalities of  all tools.
 \def\checkmark{\tikz\fill[scale=0.4](0,.35) -- (.25,0) -- (1,.7) -- (.25,.15) -- cycle;} 
\newcommand{\xmark}{\ding{55}}%
\begin{table}
    \centering
 \caption{Overview of the capabilities of the pBN synthesis tools considered.}
   \begin{tabular}{|l||*{3}{c|}}\hline
&\makebox[4em]{SamIam}&\makebox[5em]{Bayesserver}&\makebox[8em]{Storm-Prophesy}\\\hline\hline
computing sensitivity function & \xmark &$p_{\leq2}c_{\leq 2}r_1$& $p_*c_*r_*$ \\\hline
computing sensitivity value & \xmark & $p_{\leq2}c_{\leq 2}r_1$ & $p_1c_*r_*$ \\\hline
simple parameter tuning &$p_*c_1r_1$ & $p_1c_1r_1$ & $p_*c_*r_*$ \\\hline
difference parameter tuning &$p_*c_1r_1$ & $p_1c_1r_1$ & $p_*c_*r_*$ \\\hline
ratio parameter tuning &$p_*c_1r_1$& $p_1c_1r_1$& $p_*c_*r_*$ \\\hline
minimal change tuning & $p_*c_1r_1$ & $p_1c_1r_1$&  $p_*c_*r_*$ \\\hline
\end{tabular}  
    \label{tab:tools-features}
\end{table}
\\[1ex]
\noindent
\emph{Experimental set-up.}
We took benchmarks from \cite{bnlearnrepository} and conducted all our experiments on a $2.3$ GHz Intel Core i5 processor with 16 GB of RAM. We focused on questions such as:
\begin{enumerate}
\item What is the scalability for computing sensitivity functions on pBNs?
\item What is the practical scalability for feasibility checking?
\item To what extent is PLA applicable to parameter tuning for pBNs?
\end{enumerate}
\noindent
\emph{Computing sensitivity function.}
We performed a series of experiments for computing pBN sensitivity functions using our tool-chain for the $p_*c_*r_1$ class. 
Fig.~\ref{solution-function-chart} summarizes the results.
The $x-$axis (log scale) indicates the pBN benchmarks and the $y-$axis denotes the timing in seconds. 
The numbers on the bars indicate the number of parameters in the solution functions, which is related to the number of relevant parameters identified for the given query. 
We observe that Storm scales up to $380$ parameters for very large networks such as \texttt{hailfinder}. The blue bars represent regular computations, while the orange bars indicate the impact of bisimulation minimization, a built-in reduction technique in Storm.
\footnotetext{A comparison with the other tools was not applicable, as SamIam does not explicitly offers sensitivity function computation and Bayesserver sensitivity analysis is limited to 1 or 2 parameters, see Tab. \ref{tab:tools-features}.}
 \\[1ex]
\emph{Feasibility checking.}
Our tool exploits Prophesy to find a parameter instantiation $u$ of pBN $\B$ such that the BN $\B[u]$ satisfies the given inference query.
We have performed a set of experiments for the class $p_*c_*r_1$.
Fig.~\ref{feasibility-charts} (log-log scale) illustrates the results; the $x$-axis indicates the number of parameters in the pBN and the $y$-axis the time for feasibility checking (in seconds).
Each line corresponds to a pBN and the points on the lines represent single experiments.  
We inserted the parameters in the predecessors of the query nodes (i.e., in $H$) to maximize their relevance.
We also imposed queries over multiple nodes at once to push the boundaries. 
We used convex optimization (QCQP\footnote{Quadratically-constrained quadratic programming.}) (left plot) and 
PSO (right plot). 
Prophesy was able to handle up to $853$ parameters. \\[1ex]
\noindent
\emph{Approximate parameter synthesis on pBNs: Tuning the parameters and more. }
Experiments on the pBN benchmarks using PLA aimed at (a) the classes $p_1c_1r_1$ and $p_*c_1r_1$ to validate them against SamIam and Bayesserver, and (b) the class $p_*c_*r_*$ to investigate the potential of PLA for general pBNs. 
Fig.~\ref{alarm-pla-2d} visualizes the results for the \texttt{alarm} pBN with 2 parameters occurring in 26 rows of 3 CPTs, i.e., a pBN with parameter dependencies. 
The parameter $x$ was used in the CPT entries $e_1,\,\cdots,\,e_k$ only when the probability values of those entries coincided in the original BN. 
As seen in the figure, PLA can partition the entire $n$-way parameter space. 
The minimal-change parameter values can be extracted from the PLA results, where the precision depends on the PLA approximation factor.

\section{Conclusion}
\label{sec-conclusion}
This paper exploited tools and techniques for parameter synthesis on Markov chains to synthesis problems on parametric Bayesian networks. 
Prototypical tool support for pBN analysis on top of existing pMC synthesis tools has been realized. 
Our experiments indicate that pMC techniques can scale sensitivity analysis and parameter tuning tasks on pBNs. 
The experiments reveal the potential of parameter lifting~\cite{DBLP:conf/atva/QuatmannD0JK16} for partitioning the parameter space of pBNs.
Most importantly, the proposed techniques are applicable to \emph{general} pBNs --- no restrictions are imposed on the number or occurrence of parameters --- and may involve parameter dependencies. 
Future work include finding optimal parameter settings~\cite{DBLP:conf/tacas/SpelJK21}, exploiting monotonicity checking~\cite{DBLP:conf/atva/SpelJK19} and to extend the current work to (parametric) dynamic, Gaussian~\cite{DBLP:journals/aei/CastilloGHS97}, and recursive BNs~\cite{DBLP:journals/amai/Jaeger01}. 
{\paragraph*{Acknowledgement.}
We thank Robin Drahovsky for his contributions on transforming pMCs into pBNs, Caroline Jabs for her implementation efforts, and Sebastian Junges, Tim Quatmann, and Matthias Volk for discussions. We also thank Arthur Choi for his support.

\bibliographystyle{splncs04}
\bibliography{main}

\clearpage \appendix

\newpage

\section{Inference query correspondence}
\label{app:proofs} 
Here, we provide the proofs for Proposition \ref{eq:query-correspondence} and Proposition \ref{proposition-ev-tailored}. The results are formalized for non-parametric MCs of BNs and carry over to pMCs and pBNs in a straightforward manner.
\paragraph{Preliminaries and notations. }Let $\M = (S, s_0, P)$ be a MC, $\mbox{\it Paths}(\M)$ be the set of paths in $\M$ that start in the initial state $s_0$, and $\mbox{\it Paths}^*(\M)$ the set of their finite prefixes. For the path $\pi = s_0\ldots s_n \in \mbox{\it Paths}^*(\M)$, let $P(\pi) = \prod\limits_{i{=}0}^{n{-}1} P(s_i,s_{i{+}1})$. For infinite paths, a probability measure $\Pr$ is defined using a standard cylinder set construction~\cite[Ch.~10]{BK08}. Let $pre(s)$ be the direct predecessors of $s$ and $pre^*(s)$ be its reflexive and transitive closure. Let $G \subseteq S$ be the set of goal states in $\M$ and $\lozenge G$ denote the set of paths in $\M$ that eventually reach $G$, i.e., $\lozenge G = \{\pi \in \mbox{\it Paths}(\M) \, | \, \exists i \in \mathbb{N} \,.\, \pi(i) \in G \}$. Let $\mbox{\it Paths}(\M, \lozenge G) = \mbox{\it Paths}(\M) \cap (S \setminus G)^*G$, and
\begin{align}
\Pr_{\M}(\lozenge G)  ~=~ & \sum\limits_{s_0 \ldots s_n \in \mbox{\it \scriptsize Paths}(\M, \lozenge G)} P(s_0\ldots s_n).
\end{align}
For each MC, $\Pr(\lozenge G)$ is measurable. Let variable $p_{s} = \Pr_{\M}(s \models \lozenge G)$ denote the probability to reach $G$ from state $s$. It follows that
\begin{itemize}
\item if $G$ is not reachable from $s$, then $p_{s} = 0$,
\item if $s \in G$, then $p_{s} = 1$, and
\item For any state $s \in pre^*(G) \setminus G$,
\begin{align}
\label{eq:equation-system}
p_{s} = \underbrace{\sum\limits_{t \in S \setminus G} P(s,t) \cdot p_t}_{\text{reaching $G$ via state $t$}} + \underbrace{\sum\limits_{u \in G} P(s,u)}_{\text{reaching $G$ in one step}}.
\end{align}
\end{itemize}
\noindent
This yields a system of linear equations, by which $\Pr_{\M} (\lozenge G) = p_{s_0}$. \\[1ex]
Let $\square G$ denote the set of paths in $\M$ that always satisfy $G$, i.e.,
\begin{align}
\label{eq:always:eventuell:events}
& \square G ~\equiv~ \neg \lozenge \neg G \text{. Thus } 
\end{align}
\begin{align}
\label{eq:always:eventuell}
& \Pr(\square G) ~=~ 1 - \Pr(\lozenge \neg G).
\end{align}
 \noindent
 Let BN $\B = (V, W, \Theta)$ and the topological order $\varrho = (v_1,\cdots,v_m)$ on $V$ be given. Let $E$ be the evidence in the form of $E =  E_1 \wedge \ldots \wedge E_k = (v_{E_1} = d_{E_1}) \wedge  \ldots \wedge (v_{E_k} = d_{E_k})$ with $v_{E_i} <_{\varrho} v_{E_{i{+1}}}$. Recall that we often denote $E_k$ as $E_{\mbox{\it \scriptsize last}}$. 
 Let $\M^{\varrho}_{\B} = (S, s_0, P)$ be the evidence-agnostic MC of BN $\B$ as defined in Def.~\ref{defRMC}. 
 Let ${\M^\varrho_{\B,E}}$ denote the evidence-tailored MC of the BN $\B$. 
\paragraph{Query correspondence for the evidence-agnostic MC. }
 We prove Proposition \ref{eq:query-correspondence} by indicating the strong correspondence between the paths of the evidence-agnostic MC of $\B$ and the tree-like MC \cite{DBLP:conf/qest/SalmaniK20} of $\B$.  Let $\boldsymbol{\M^{\varrho}_\B} = (\boldsymbol{S}, \boldsymbol{s_0}, \boldsymbol{P})$ be the tree-like MC of $\B$ given the topological order $\varrho$. 
\noindent The construction of $\boldsymbol{\M^{\varrho}_\B}$ \cite{DBLP:conf/qest/SalmaniK20} is similar to that of $\M^{\varrho}_{\B}$ and only deviates in the following. Let $\boldsymbol{s_{i-1}}$ be a state at the $i{-}1$'th level of $\boldsymbol{\M^{\varrho}_\B}$. The successor states of $\boldsymbol{s_{i-1}}$ are obtained for each $d_i \in D_{v_i}$ by $\boldsymbol{s_{i-1}}[v_i = d_i] = \boldsymbol{s}$, where
\begin{itemize}
\item $\boldsymbol{s}(v_i) = d_i$ and
\item $\boldsymbol{s}(v) = \boldsymbol{s_{i{-}1}}(v)$ for $v \neq v_i$.
\end{itemize}
The transition probabilities are determined analogously to that of $\M^{\varrho}_\B$. This tree-like MC is thus obtained by considering all possible valuations of the variables in BN $\B$ adhering to the topological order $\varrho$. Inference on the BN $\B$ can be reduced to computing reachability probabilities on the tree-like MC $\boldsymbol{\M^{\varrho}_\B}$, as
\begin{align}
\label{eq:tree-like}
  \Pr_{\B}(E) ~=~ \Pr_{\boldsymbol{\M^{\varrho}_\B}}(\lozenge \, E) \quad \text{ and } \quad \Pr_{\B}(H \wedge E) ~=~ \Pr_{\boldsymbol{\M^{\varrho}_\B}}(\lozenge \, (H \wedge E)).
\end{align}
This is formally shown in \cite{DBLP:conf/qest/SalmaniK20}. \\ [1ex]
Let $A \subseteq V$ be a subset of BN variables and $\eta \in Eval(A)$ be an evaluation of variables in $A$. Let $\mbox{\it Paths}(\boldsymbol{\M^{\varrho}_{\B}}, \lozenge \,\eta) = \boldsymbol{s_0s_1} \ldots \in \mbox{\it Paths}(\boldsymbol{\M^{\varrho}_{\B}})$ such that 
$\exists \boldsymbol{s_k}  \text{ with } \boldsymbol{s_k}(v){=}\eta(v) \text{ for all } v \in A$. Let $\mbox{\it Paths}(\M^{\varrho}_{\B}, \square \, \eta^*) = s_0s_1 \ldots \in \mbox{\it Paths}({\M^{\varrho}_{\B}})$ such that $\text{for each } s_i \text{ and each } v{\in}A, s_i(v){=}\eta(v) \text{ or } s_i(v){=}*$. 
\begin{lemma}
\label{lemma:1}
 Path $\pi = s_0 s_1 \ldots \in \mbox{\it Paths}(\M^{\varrho}_{\B}, \square \, \eta^*)$ if and only if there is the corresponding path $\boldsymbol{\pi} = \boldsymbol{s_0} \boldsymbol{s_1} \ldots \in \mbox{\it Paths}(\boldsymbol{\M^{\varrho}_{\B}}, \lozenge \eta)$ such that for each $i \in \mathbb{N}$,
\[
 P(s_i, s_{i+1}) = \boldsymbol{P}(\boldsymbol{s_i}, \boldsymbol{s_{i+1}}).
\]
\noindent
\end{lemma}
\vspace{-0.3cm}
\noindent
\paragraph{Proof. }We indicate the correspondence between the paths in one direction. The proof is analogous for the other direction. Let $\boldsymbol{\pi} =  \boldsymbol{s_0} \boldsymbol{s_1} \ldots \in \mbox{\it Paths}(\boldsymbol{\M^{\varrho}_{\B}}, \lozenge \eta)$. 
By the definition of $\boldsymbol{\M^{\varrho}_{\B}}$, 
\begin{itemize}
\item $\boldsymbol{s_{0}}$ is the initial state,
\item for $1 \leq i \leq m$, $\boldsymbol{s_{i}} \in \boldsymbol{S_{i}}$ is state at the $i$'th level of $\boldsymbol{\M^{\varrho}_{\B}}$, and
\item for each $\boldsymbol{s_{i}} \in \boldsymbol{S_{i}} $, $\exists!\, d_{i} \in D_{v_i}$ such that $\boldsymbol{s_i} \models  (v_{i} = d_{i})$\footnote{Note that $v_{i}$ is the $i$'th variable in the topological order $\varrho$.}.
\end{itemize}
 Let $\mbox{\sl vals}_{\boldsymbol{\pi}}= d_1d_2\cdots d_m$ be the sequence of the values for BN variables along the path $\boldsymbol{\pi}$, where $\mbox{\sl vals}_{\boldsymbol{\pi}}(i) = d_i \in D_{v_i}$ iff $\boldsymbol{s_i} \models (v_i = d_i)$.
 The corresponding path ${\pi} \in \mbox{\it Paths}({\M^{\varrho}_{\B}}, \square \eta^*)$ is then obtained as follows. \\ [1ex]
Let $0 \leq i < m$ and $s_{i} \in S_{i}$ be a state at the ${i}$'th level of ${\M^{\varrho}_{\B}}$. Def. \ref{defRMC} ensures that
\begin{center}
 - for each $d_{i{+}1} \in D_{v_{i{+}1}} $, $\exists!\, s_{i{+}1} \in \mbox{\sl succ}({s_{i}})$ such that ${s_{i{+}1}} \models  (v_{i{+}1} = d_{i{+}1})$.
 \end{center}
  Given $\mbox{\sl vals}_{\boldsymbol{\pi}}$, the path $\pi = s_0s_1\cdots \in  \mbox{\it Paths}(\M^{\varrho}_{\B})$ is then obtained, where
\begin{itemize}
\item $s_0$ is the initial state of ${\M^{\varrho}_{\B}}$,
\item for $0 \leq i < m$, $s_{i{+}1}$ is the unique state in $\mbox{\sl succ}({s_{i}})$ with
\begin{center} ${s_{i{+}1}} \models  (v_{i{+}1} = \mbox{\sl vals}_{\boldsymbol{\pi}}(i{+}1))$,
\end{center}
\item and for $i \geq m$, $s_{i{+}1} = s_m$.
\end{itemize}
Note that for $0 \leq i < m$, $P(s_{i}, s_{i{+}1}) = \boldsymbol{P}(\boldsymbol{s_{i}}, \boldsymbol{s_{i{+}1}}) = \Theta(\overline{par})(\mbox{\sl vals}_{\boldsymbol{\pi}}(i{+}1)) $ for the parent valuation $\overline{par} \in \mbox{\it Eval}(\mbox{\it parents}(v_{i{+}1}))$ with $s_{i} \models \overline{par}.$ For $i \geq m$, $P(s_{i}, s_{i{+}1}) = \boldsymbol{P}(\boldsymbol{s_{i}}, \boldsymbol{s_{i{+}1}}) = 1$. \\ [1ex]
It remains to show that  ${\pi} \in \mbox{\it Paths}({\M^{\varrho}_{\B}}, \square \eta^*)$: \\ [1ex]
(a) Path $\boldsymbol{\pi} = \boldsymbol{s_0}\boldsymbol{s_1}\cdots  \in \mbox{\it Paths}(\boldsymbol{\M^{\varrho}_{\B}}, \lozenge \eta)$. Thus, $\exists k$ such that $\boldsymbol{s_k}(v) = \eta(v)$ for all the variables $v \in A$. \\ [1ex]
(b) The BN variable $v$ is not assigned to two distinct values in $D_v$ along a given path in $\mbox{\it Paths}(\boldsymbol{\M^{\varrho}_{\B}})$, i.e.,  for $d \in D_v$,
\begin{align*}
\boldsymbol{s_k}(v) = d \implies \boldsymbol{s_i}(v) = d \lor \boldsymbol{s_i}(v) = * \text{ for } i < k \text{ and } i > k. 
\end{align*}
(c) Let $\boldsymbol{s_i} \in \boldsymbol{S_i}$ be a state at the $i$'th level of $\boldsymbol{\M^{\varrho}_{\B}}$ and $v_i$ be the $i$'th variable in the topological order. Then by the definition of $\boldsymbol{\M^{\varrho}_{\B}}$, $\boldsymbol{s_i}(v_i) \neq *$. \\ [1ex]
 It follows by (a), (b), and (c) that 
  $s_i(v_i) = \mbox{\sl vals}_{\boldsymbol{\pi}}(i) = \eta(v_i)$ for $v_i \in A$. Moreover, similarly to argument (b), it holds that for $d_i \in D_{v_i}$,
 \begin{align*}
 {s_i}(v_i) = d_i \implies {s_j}(v_i) = d_i \lor {s_j}(v_i) = * \text{ for } j < i \text{ and } j > i. 
\end{align*}
 It thus yields ${\pi} \in \mbox{\it Paths}({\M^{\varrho}_{\B}}, \square \eta^*)$. This ends the proof for Lemma \ref{lemma:1}. \hfill $\boxtimes$ \\ [1ex]
For the evidence $E = \bigwedge\limits_{i = 1}^{k}(v_i = d_i )$, let $E^* = \bigwedge\limits_{i = 1}^{k}(v_i = d_i \lor v_i = *)$.\footnote{$H^*$ is defined analogously.} The following equations are obtained by considering $\eta := E$ and $\eta := E \wedge H$ in Lemma \ref{lemma:1}.
\begin{align}
\label{lemma1:correspondence-eq}
 \Pr_{\boldsymbol{\M^{\varrho}_\B}}(\lozenge \, E) ~=~ \Pr_{\M^{\varrho}_{\B}}(\square \,  E^*) \quad \text{ and }  \quad\Pr_{\boldsymbol{\M^{\varrho}_\B}}(\lozenge \, (H \wedge E)) ~=~ & \Pr_{\M^{\varrho}_{\B}}(\square \,  (H^* \wedge E^*)).
\end{align}
\noindent
\textbf{Proposition 1. }For the pMC $\M^{\varrho}_{\B}$,
\[
\Pr_{\B}(E) \ = \ 1 - \Pr_{\M^{\varrho}_{\B}}(\lozenge \, \neg E) 
\quad \text{and} \quad
\Pr_{\B} (H \mid E) \ = \ \dfrac{1 - \Pr_{\M^\varrho_\B}(\lozenge\,(\neg H \lor \neg E))}{1 - \Pr_{\M^{\varrho}_{\B}}(\lozenge \, \neg E)},
\]
where the latter equality requires $\Pr_{\B}(\neg E) < 1$.
\paragraph{Proof. } It follows from equations (\ref{eq:tree-like}) and (\ref{lemma1:correspondence-eq}) that
    \begin{align}
    \label{eq:query-correspondence-evidence}
     \Pr_{\B}(E) ~=~ \Pr_{\M^{\varrho}_{\B}}(\square \,  E^*) \quad \text{ and } \Pr_{\B}(H \wedge E) ~=~ \Pr_{\M^{\varrho}_{\B}}(\square \,  (H^* \wedge E^*)).
     \end{align}
     Without loss of generality, assume that the BN variables in $V$ are binary-valued with $D_{v_i} = \{d_i, \neg d_i\}$. 
 Definition \ref{defRMC} ensures that for all $s \in S$, $s \models (v_i = d_i) \lor (v_i = \neg d_i) \lor (v_i = *)$. Let $\neg E = \bigvee_{i = 1}^{k}(v_i = \neg d_i)$. It follows that
 \begin{align*}
 \neg (E^*) ~\equiv~ \neg E, \quad \text{thus}
 \end{align*}
\begin{align}
\label{eq:neg:E-star}
 \Pr_{\M^{\varrho}_{\B}}(\square \,  E^*) ~=~ 1 - \Pr_{\M^{\varrho}_{\B}}(\lozenge \, \neg (E^*)) ~=~  1 - \Pr_{\M^{\varrho}_{\B}}(\lozenge \, \neg E).
\end{align}
Since $E$ allows the conjunction of multiple variables, equation (\ref{eq:neg:E-star}) is analogously valid for $(E \wedge H)$, i.e.,
\begin{align}
\label{eq:neg:E-H-star}
\Pr_{\M^{\varrho}_{\B}}(\square \, (H^* \wedge E^*)) ~=~ 1 - \Pr_{\M^{\varrho}_{\B}}(\lozenge \, (\neg H \lor \neg E)).
\end{align}
We can then derive:
\begin{align*}
       & \Pr_{\B} (H \mid E) \\
       ~=~ &  \dfrac{\Pr_{\B} (H \wedge E)}{\Pr_{\B} (E)}\\
       \stackrel{(\ref{eq:query-correspondence-evidence})}{=}~ & \dfrac{\Pr_{\M^\varrho_\B}\big(\square\, (H^* \wedge E^*)\big)}{\Pr_{\M^{\varrho}_{\B}}(\square E^*)} \\
     \stackrel{(\ref{eq:neg:E-star}) \text{ and }(\ref{eq:neg:E-H-star})}{=}~ & \dfrac{1 - \Pr_{\M^\varrho_\B}(\lozenge\,(\neg H \lor \neg E))}{1 - \Pr_{\M^{\varrho}_{\B}}(\lozenge \, \neg E)}.
\end{align*}
This finalizes the proof for Proposition \ref{eq:query-correspondence}. \hfill $\boxtimes$
\paragraph{Formalizing the operations. }
We now proceed towards the proof of Proposition \ref{proposition-ev-tailored} by first formalizing the two operations to construct the \emph{evidence-tailored} MC of BN: \emph{propagation} and \emph{redirection}.

\begin{definition}[Propagation]
\label{prop:def}
Let $\M^{\varrho}_{\B}$ be the evidence-agnostic MC of BN $\B$ for the topological order $\varrho$. Applying the propagation operation on $\M^{\varrho}_{\B}$ yields the MC $\M^{\varrho\downarrow}_{\B} = (S^{\downarrow}, s^{\downarrow}_0, P^{\downarrow})$, where
\begin{itemize}[label={--}]
\item $s^{\downarrow}_0 = s_0$ is the initial state.
\item $S^{\downarrow} = \bigcup\limits_{i{=}0}^{m} S^{\downarrow}_i$ is the set of states defined as follows:
\begin{itemize}
\item For $i{=}0$ and $i > \varrho(v_{\Elast})$, \\
$S^{\downarrow}_i = S_i$,  where $S_i$ is the set of states at level $i$ of $\M^{\varrho}_{\B}$.
\vspace{1ex}
\noindent
\item For $0 < i \leq \varrho_{\Elast}$, \\
$S^{\downarrow}_i = \{\underbrace{s^{\downarrow}_{i{-}1}[v_i{=}d_i]}_{\text{short } s^{\downarrow}} \,\mid\, \text{for } s^{\downarrow}_{i{-}1} \in S^{\downarrow}_{i{-}1} \text{ and } d_i \in D_{v_i}\}, \text{ with }$
\vspace{1ex}
\begin{itemize}[label={$\bullet$}]
\item $s^{\downarrow}(v_i) = d_i$,
\item $s^{\downarrow}(v) = s^{\downarrow}_{i{-}1}(v)$ for $v \neq v_i$ iff either
\vspace{1ex}
\begin{center}
\begin{enumerate}
\item[(I)]  $\varrho(v) < \varrho(v_i) \text{ and } \exists c \in children(v) \,.\, \varrho(c) > \varrho(v_i)$ or
\vspace{1ex}
\item[(II)] $v \not \in \mbox{\it vars}(E)$,
\end{enumerate}
\end{center}
\vspace{1ex}
\item and $s^{\downarrow}(v) = *$, otherwise.
\end{itemize}
\end{itemize}
\vspace{1ex}
 \item $P^{\downarrow}$ is defined analogously to $P$.
 \end{itemize}
 \end{definition}
 The MC $\M^{\varrho\downarrow}_{\B}$ differs from $\M^{\varrho}_{\B}$ for $0 < i \leq \varrho(v_{\Elast})$ as imposed by constraint (II).
 \begin{lemma}
 \label{propagation:lemma}
  The propagation operation ensures that
 \begin{align}
 \label{propagation:equation}
  & \Pr_{\B}(H | E) ~=~ \dfrac{\Pr_{\M^{\varrho\downarrow}_{\B}}( \square  (H^* \wedge E^*))}{\Pr_{\M^{\varrho\downarrow}_{\B}}( \square E^*)}. 
 \end{align}
 \end{lemma}
 \paragraph{Proof}\footnote{The proof is analogous to Lemma \ref{lemma:1}.}. 
  Definition \ref{prop:def} ensures a strong correspondence between the paths in $\M_{\B}^{\varrho}$ and $\M^{\varrho\downarrow}_{\B}$: $\pi^{\downarrow}= s^{\downarrow}_0s^{\downarrow}_1\ldots \in \mbox{\it Paths}(\M^{\varrho\downarrow}_{\B}, \lozenge E)$ if and only if there is the unique path $\pi = s_0s_1\ldots \in \mbox{\it Paths}(\M_{\B}^{\varrho}, \square E^*)$ such that for each $i \in \mathbb{N}$,
 \begin{align*}
 P^{\downarrow}(s^{\downarrow}_i, s^{\downarrow}_{i{+}1}) = P(s_i,s_{i{+}1}).
 \end{align*}
 It follows that
   \begin{align}
   \label{propagation:numinator}
 & \Pr_{\M_{\B}^{\varrho}}(\square E^*) ~=~ \Pr_{\M^{\varrho\downarrow}_{\B}}(\square E^*) 
 \end{align}
 and analogously for $H \wedge E$,
 \begin{align}
    \label{propagation:denuminator}
   &  \Pr_{\M_{\B}^{\varrho}}(\square ( H^* \wedge E^*)) ~=~ \Pr_{\M^{\varrho\downarrow}_{\B}}(\square (  H^* \wedge E^*)). 
   \end{align}
   Equation (\ref{propagation:equation}) derives from the equations (\ref{propagation:numinator}) and (\ref{propagation:denuminator}), and Proposition \ref{eq:query-correspondence}. \hfill $\boxtimes$\\ [2ex]
\noindent
 We now proceed by formalizing the redirection operation for an arbitrary MC. Let $\M = (S, s_0, P)$ be a MC and $AP = \{a_i\,\, | \, i \in \mathbb{N}\}$ be a set of atomic propositions. Let $L : S \to 2^{AP}$ be a labelling function that maps each state of  $\M$ to a finite subset of atomic propositions. Let $\phi = \bigwedge\limits_{i=1}^{k} a_i$ and $S_{\neg \phi}$ denote the set of states that violate $\phi$, i.e., 
 \begin{align*}
 S_{\neg \phi} = \{s \in S \,|\, s \models \neg a_i \text{ for some } 1 \leq i \leq k \}.
 \end{align*}
  The \emph{redirection operation} reroutes the direct transitions to the states in $S_{\neg \phi}$ to the initial state $s_0$ and deletes the states in $S_{\neg \phi}$.
 \begin{definition}[Redirection]
 \label{redirection-definition} Applying the redirection operation to the MC $\M$ with respect to $\phi$ yields the MC $(\M)_{\phi} = (S \setminus S_{ \neg \phi}, s_0, P_{\phi})$, where for $s \in S \setminus S_{ \neg \phi}$
 \[
 \left\{
	\begin{array}{ll}
		P_{\phi}(s, s_0) = P(s, s_0) ~+~ \mathlarger{\sum}\limits_{s_{\neg} \in S_{\neg \phi}}P(s, s_{\neg}) &  \quad \text{ and} \\
		[3ex]
		P_{\phi}(s, s') = P(s, s' )&  \quad \text{ for } s' \in  S \setminus S_{ \neg \phi} \text{ with } s' \neq s_0.
	\end{array}
	    \right.
  \]
  \end{definition}
    \noindent
 Let $G$ be a set of target states in $(\M)_{\phi}$. Recall that variable  $p_{s}$ denotes the probability to reach $G$ from state $s$. Let $\mbox{\sl succ}(s)$ denote the set of direct successors of $s$ in the MC $\M$. 
It follows by equation (\ref{eq:equation-system}) and Def. \ref{redirection-definition} that for the state $s \in S \setminus S_{\neg \phi}$,
 \begin{align}
 \label{equation:system:tailored}
& p_{s} ~=~ \mathlarger{\sum}\limits_{s' \in  \mbox{\sl \scriptsize succ}(s) \setminus S_{\neg \phi}}P(s, s') \cdot p_{s'} ~+~ \underbrace{\mathlarger{\sum}\limits_{s_{\neg} \in  \mbox{\sl \scriptsize succ}(s) \cap S_{\neg \phi}} P(s, s_{\neg}) \cdot p_{s_0}.}_{\text{ redirection to the initial state }}
\end{align}
           \begin{figure}[t]
           \centering
              \includegraphics[scale=0.1]{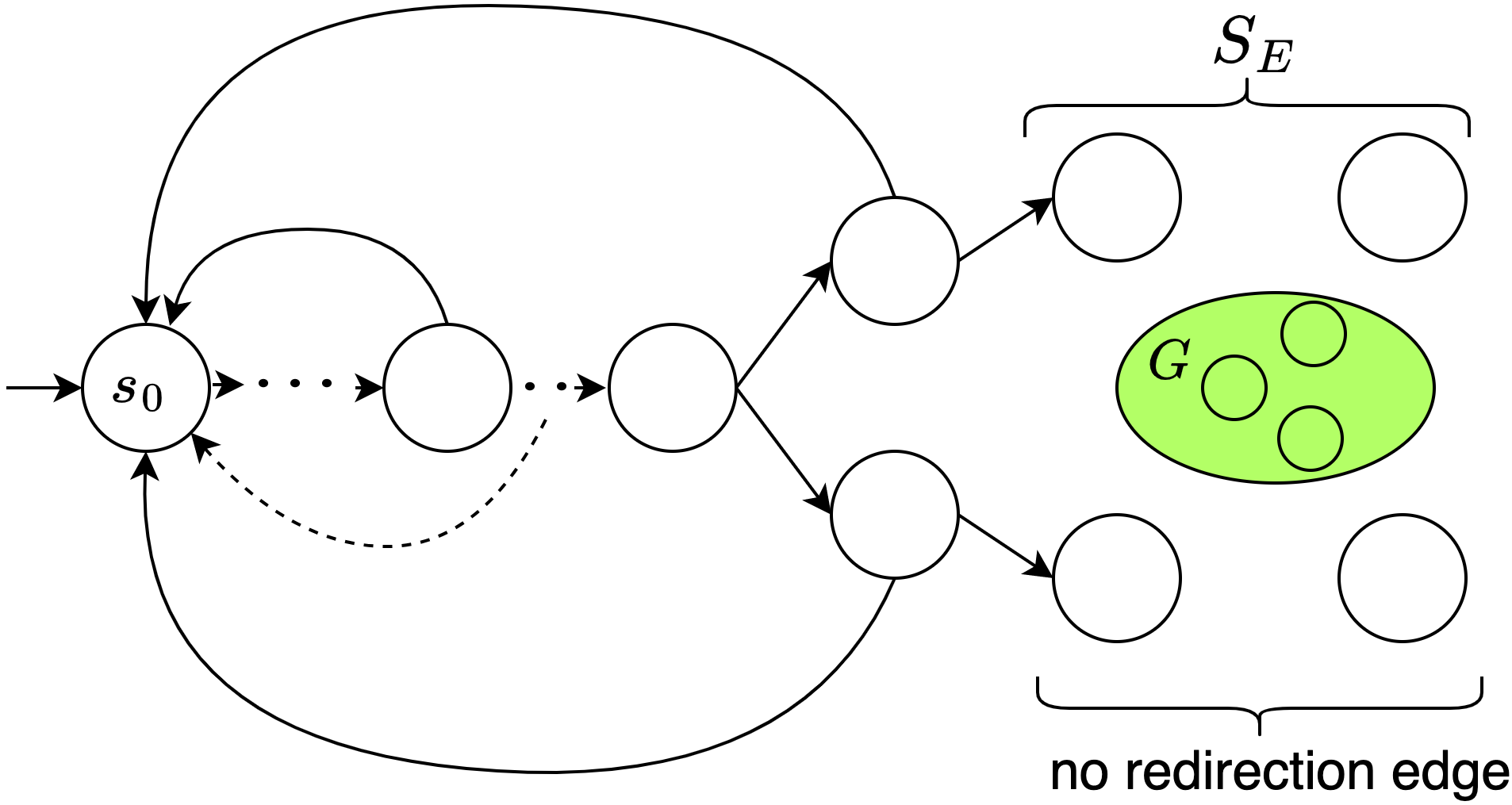}
              \caption{The goal states $G$ occur after the last redirection loop of $\M_{\B, E}$: $G \subseteq S_E$.}
                     \label{fig:proof:sketch}
       \end{figure}   
\noindent 
\paragraph{Query correspondence for the evidence-tailored MC.}
The evidence-tailored MC $\M^{\varrho}_{\B,E} = (S^{\downarrow} \setminus S^{\downarrow}_{\neg E}, s^{\downarrow}_0, P_E)$ is obtained by applying the redirection operation to the MC $\M^{\varrho\downarrow}_{\B}$ with respect to $\phi = E$, i.e., $\M^{\varrho}_{\B,E} = (\M^{\varrho\downarrow}_{\B})_E$. Let $S_E = \bigcup\limits_{i=\varrho(v_{E_{\tiny last}})}^{m}S^{\downarrow}_{i}$ be the set of states in $\M^{\varrho}_{\B,E}$ that only occurs at the level of the last evidence node and afterwards, i.e., the set of states that occur after the last redirection edge, see Fig. \ref{fig:proof:sketch}.
 \begin{lemma}
\label{lemma:redirection-devision}
  For $G \subseteq S_E$, it holds that
\begin{align*}
& \Pr_{{\M^\varrho_{\B,E}}}(\lozenge G)  ~=~  \dfrac{\Pr_{\M^{\varrho\downarrow}_{\B}}(\lozenge G \wedge \square E^*)}{\Pr_{\M^{\varrho\downarrow}_{\B}}(\square E^*)}.
\end{align*}
\end{lemma}
\paragraph{Proof. }We prove the above lemma by rewriting equation (\ref{equation:system:tailored}) for $\M^{\varrho}_{B, E}$. First, let $G \subseteq S^{\downarrow}$ be arbitrary. Recall that $S^{\downarrow}_i$ denotes the set of states at the $i$-th level of  $\M^{\varrho\downarrow}_{\B}$.\footnote{Note that by Def. \ref{prop:def}, for every $s^{\downarrow}_i \in S^{\downarrow}_i$, $\mbox{\sl succ}(s^{\downarrow}_i) \subseteq S^{\downarrow}_{i{+}1}$.}
\begin{itemize}
\item For each $s^{\downarrow}_i \in S^{\downarrow}_i \cap pre^*(G)$,
\begin{align*}
 p_{s^{\downarrow}_i} = \mathlarger{\sum}\limits_{s^{\downarrow}_{i{+}1} \in S^{\downarrow}_{i{+}1} \setminus S^{\downarrow}_{\neg E} } P^{\downarrow}(s^{\downarrow}_i, s^{\downarrow}_{i{+}1}) \cdot p_{s^{\downarrow}_{i{+}1}} ~+~ \mathlarger{\sum}\limits_{t^{\downarrow}_{i{+}1}\in S^{\downarrow}_{i{+}1} \cap S^{\downarrow}_{\neg E} } P^{\downarrow}(s^{\downarrow}_i, t^{\downarrow}_{i{+}1}) \cdot p_{s^{\downarrow}_{0}},
\end{align*}
\item for each $s^{\downarrow}_i \not \in pre^*(G)$, $p_{s^{\downarrow}_i} = 0$,
\item and for each $s^{\downarrow}_i \in G$, $p_{s^{\downarrow}_i} = 1$.
\end{itemize}
 \noindent
 Let $\neg E_{1 \ldots j} = \bigcup\limits_{i{=}1}^{\varrho(v_{E_j})} S^{\downarrow}_i \cap S^{\downarrow}_{\neg E}$ specify the states that violate $E_1$ up to $E_j$. The above equation system yields the following. \\ [1ex]
 \noindent For $G \subseteq \bigcup\limits_{i = \varrho(v_{E_j})}^m S^{\downarrow}_i$,
 \begin{align*}
 & p_{s^{\downarrow}_{0}} \quad \quad ~=~  \quad \quad \mathlarger{\sum}\limits_{\mathclap{\footnotesize \pi \in \mbox{\it \scriptsize Paths}\big(\M^{\varrho\downarrow}_{\B}, \lozenge G \big) \setminus \mbox{\it \scriptsize Paths}\big(\M^{\varrho\downarrow}_{\B}, \lozenge \neg E_{1\ldots j}\big)}} ~\Pr(\pi) \quad \quad \quad ~~+~ \quad \quad \quad p_{s^{\downarrow}_{0}} ~\cdot~ \mathlarger{\sum}\limits_{\mathclap{\footnotesize \pi' \in \mbox{\it \scriptsize Paths}\big(\M^{\varrho\downarrow}_{\B}, \lozenge \neg E_{1 \ldots j}\big)}} ~\Pr(\pi')~
 \end{align*}
and for $G \subseteq  \bigcup\limits_{i = \varrho(v_{E_{\tiny last}})}^m S^{\downarrow}_i$,
\begin{align}
\label{nice-eq}
&  p_{s^{\downarrow}_{0}}  \quad \quad ~=~ \quad \quad   \mathlarger{\sum}\limits_{\mathclap{\footnotesize \pi \in \mbox{\it \scriptsize Paths}\big(\M^{\varrho\downarrow}_{\B}, \lozenge G\big) \setminus \mbox{\it \scriptsize Paths}\big(\M^{\varrho\downarrow}_{\B}, \lozenge \neg E\big)}} ~\Pr(\pi)~  \quad \quad \quad ~+~\quad \quad \quad p_{s^{\downarrow}_{0}} ~\cdot~ \mathlarger{\sum}\limits_{\mathclap{\footnotesize \pi' \in \mbox{\it \scriptsize Paths}\big(\M^{\varrho\downarrow}_{\B}, \lozenge \neg E\big)}} ~\Pr(\pi')~.
\end{align}
We have $\neg (\lozenge \neg E) \equiv \square E^*$ (Equations (\ref{eq:always:eventuell:events}) and (\ref{eq:neg:E-star})). Moreover, note that $A \setminus B = A \cap \neg B$. We can thus rewrite equation (\ref{nice-eq}) as
\begin{align*}
& \Pr_{\M^\varrho_{\B,E}}(\lozenge G) ~=~ p_{s^{\downarrow}_0} ~=~ {\Pr_{\M^{\varrho\downarrow}_{\B}}(\lozenge G \wedge \square E^*)} ~+~ p_{s^{\downarrow}_0} \cdot {\Pr_{\M^{\varrho\downarrow}_{\B}}(\lozenge \neg E)} .
\end{align*}
Thus:
\begin{align*}
&p_{s^{\downarrow}_0} ~=~  \Pr_{{\M^\varrho_{\B,E}}}(\lozenge G)  ~=~  \dfrac{\Pr_{\M^{\varrho\downarrow}_{\B}}(\lozenge G \wedge \square E^*)}{1 - \Pr_{\M^{\varrho\downarrow}_{\B}}(\lozenge \neg E)}.
\end{align*}
 Note that ${1 - \Pr_{\M^{\varrho\downarrow}_{\B}}(\lozenge \neg E)}  = {\Pr_{\M^{\varrho\downarrow}_{\B}}(\square E^*)}$. This ends the proof. \hfill $\boxtimes$ \\ [1ex]
\noindent
 \textbf{Proposition 2.}
For the evidence-tailored pMC $\M^{\varrho}_{\B, E}$ of pBN $\B$, we have:
\[
          \Pr_{\B}(H \mid E) \ =  \  1 - \Pr_{\M^{\varrho}_{\B, E}}\big(\lozenge \big( \, (\neg \Hbefore \wedge \Elast) \lor \neg \Hafter \big)\big).
\]

\paragraph{Proof. } We apply Lemma \ref{lemma:redirection-devision} to the evidence-tailored MC $\M^{\varrho}_{\B, E}$. Let $G = \big( \, (\neg \Hbefore \wedge \Elast) \lor \neg \Hafter )\big)$. Recall that $S_E$ is the set of states at the levels $i$ with $i \geq \varrho(v_{\Elast})$. 
We have $G \subseteq S_{E}$: the term $\neg \Hbefore \wedge \Elast$ can first be satisfied at level $i$ with $i \geq \varrho(v_{\Elast})$. The term $\Hafter$ can also first be satisfied at levels $i$ with $i > \varrho(v_{\Elast})$ by the definition of $\Hafter$. \\ [1ex]
\noindent
 It then follows by Lemma \ref{lemma:redirection-devision} that 
\begin{align*}
&\ 1 - \Pr_{\M^{\varrho}_{\B, E}}\big(\lozenge G \big) \\
~=~ &  \dfrac{\Pr_{\M^{\varrho\downarrow}_{\B}}(\square E^*\big) - \Pr_{\M^{\varrho\downarrow}_{\B}}\big(\lozenge G \wedge \square E^*\big)}{\Pr_{\M^{\varrho\downarrow}_{\B}}(\square E^*\big)} 
\end{align*}
\vspace{-1ex}
\begin{align*}
~\stackrel{\Pr(a) - \Pr(a \wedge b) = \Pr(a \wedge \neg b)}{=}~ &  \dfrac{\Pr_{\M^{\varrho\downarrow}_{\B}}\big({\square E^*} \wedge \, \neg {\lozenge G}\big)}{\Pr_{\M^{\varrho\downarrow}_{\B}}(\square E^*\big)}  \\ 
~\stackrel{\neg \lozenge a \equiv \square \neg a}{=}~ & \dfrac{\Pr_{\M^{\varrho\downarrow}_{\B}}\big({\square E^*} \wedge \, \square {\neg G}\big)}{\Pr_{\M^{\varrho\downarrow}_{\B}}(\square E^*\big)} \\ 
~\stackrel{\square \neg (a \lor b) \equiv \square \neg a \wedge \square \neg b}{=}~& \dfrac{\Pr_{\M^{\varrho\downarrow}_{\B}}\big({\square E^*} \wedge \, \square {\neg \, (\neg \Hbefore \wedge \Elast) \wedge \neg( \neg \Hafter)}\big)}{\Pr_{\M^{\varrho\downarrow}_{\B}}(\square E^*\big)}  \quad \quad \\
~\stackrel{\neg (\neg H) \equiv H^*}{=}~ & \dfrac{\Pr_{\M^{\varrho\downarrow}_{\B}}\big({\square E^*} \wedge \, \square {(\Hbefore^* \lor \neg \Elast) \wedge \square \Hafter^*}\big)}{\Pr_{\M^{\varrho\downarrow}_{\B}}(\square E^*\big)} \\ 
~\stackrel{\square a \wedge \square b \equiv\wedge \square (a \wedge b)}{=}~ & \dfrac{\Pr_{\M^{\varrho\downarrow}_{\B}}\big( \square {(E^* \wedge (\Hbefore^* \lor \neg \Elast))\wedge \square \Hafter^*}\big)}{\Pr_{\M^{\varrho\downarrow}_{\B}}(\square E^*\big)} \\ 
\end{align*}
\vspace{-1cm}
\begin{align*}
~=~ & \dfrac{\Pr_{\M^{\varrho\downarrow}_{\B}}\big( \square {( (E^* \wedge\Hbefore^*) \lor {(E^* \wedge \neg \Elast)})\wedge \square \Hafter^*}\big)}{\Pr_{\M^{\varrho\downarrow}_{\B}}(\square E^*\big)}
\end{align*}
\begin{align*}
~\stackrel{E^* \wedge \neg \Elast \equiv \texttt{false}}{=}~ & \dfrac{\Pr_{\M^{\varrho\downarrow}_{\B}}\big({\square (E^* \wedge \Hbefore^* \wedge \Hafter^*)}\big)}{\Pr_{\M^{\varrho\downarrow}_{\B}}(\square E^*\big)} \\
~\stackrel{\Hbefore^* \wedge \Hafter^* \equiv H^*}{=}~ & \dfrac{\Pr_{\M^{\varrho\downarrow}_{\B}}\big({\square (E^* \wedge H^* )}\big)}{\Pr_{\M^{\varrho\downarrow}_{\B}}(\square E^*\big)}.  
\end{align*}
It then follows by Lemma \ref{propagation:lemma} that 
\begin{align*}
 1 - \Pr_{\M^{\varrho}_{\B, E}}\big(\lozenge \big( \, (\neg \Hbefore \wedge \Elast) \lor \neg \Hafter \big)\big) ~=~ \Pr_{\B}(H \mid E),
\end{align*}
which finalizes the proof for Proposition \ref{proposition-ev-tailored}. \hfill $\boxtimes$
\section{Some detailed experimental results}
\label{app:details-par-tuning}
\newcommand*{\QEDA}{\hfill\ensuremath{\blacksquare}}
\newcommand{\init}{{\mbox{\it \footnotesize init}}}
Figure \ref{PLA-precision} and Tables \ref{sensitivity-table}, \ref{tab:one-way}, and  \ref{tab:pla-multiway} provide a detailed overview on some of our experiments for pBN parameter tuning.
\noindent
\paragraph{Storm's precision for minimal change parameter tuning. }Figure \ref{PLA-precision} provides an overview of how precise the results of \emph{parameter space partitioning} experiments are in comparison to Bayesserver, one of our baseline tools. Note that Bayesserver reports the result with higher precision in comparison to SamIam. We took a $p_1c_1d_1$ subclass from \texttt{alarm} network and performed approximate parameter tuning using our tool. Each point indicates a single experiment. The plot is in log-log scale. The x-axis indicates the approximation factor $(1{-}c)$ for the experiments; the y-axis indicates how much the results ---the minimal distance point--- found by PLA deviates from the results reported by Bayesserver. We see that the precision tightly depends on the approximation factor and as we increase PLA's coverage --- i.e., smaller approximation factor---, the results get closer to that of reported by Bayesserver. The results coincide at $1{-}c=10^{-16}$. 
\begin{figure}
\vspace*{-0.5cm}
\centering
 \includegraphics[scale=0.45]{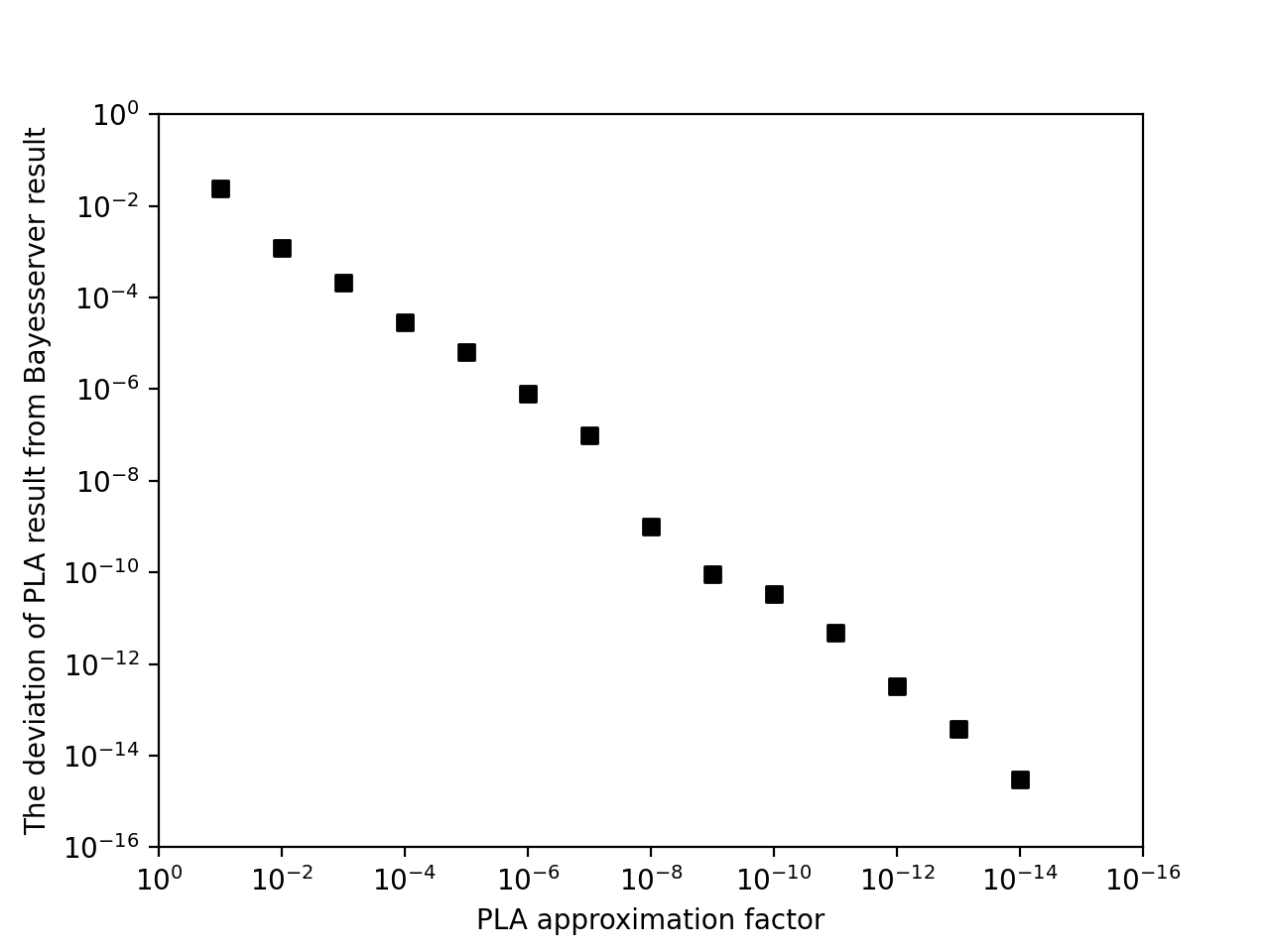}
 \vspace*{-0.3cm}
 \caption{The difference between Storm results and Bayesserver results for various PLA approximation factors, $p_1c_1d_1$ subclass, \texttt{alarm} network. }
 \label{PLA-precision}
\vspace*{-0.5cm}
\end{figure}
\paragraph{Appendix tables description.}
Table \ref{sensitivity-table} (page \pageref{sensitivity-table}) includes the detailed results for computing sensitivity function. 
For each network, three pBN versions were considered: $p_1c_1r_1$, $p_2c_{\leq}r_1$, and $p_*c_*r_1$. 
Column ``$\#$parameters'' indicates the total number of parameters. 
Column ``Storm (regular)'' indicates the model checking time taken by Storm for computing the sensitivity functions, while the column ``Storm (bi-simulation)'' indicates the sensitivity function computation time when exploiting bisimulation minimization. 
The last column indicates the corresponding sensitivity analysis time for Bayesserver, for the same pBN and query. Note that Bayesserver does not support the class $p_*c_*r_1$, thus we compared the results for the subclasses $p_1c_1r_1$ and $p_2c_{\leq}r_1$. We see that Storm (when using the bi-simulation engine) outperforms Bayesserver in most of the experiments.
\indent Table \ref{tab:one-way} (page \pageref{tab:one-way}) indicates our experimental results for parameter tuning on the $p_1c_1d_1$ class, the only subclass supported by both baseline tools. 
For Storm, the column ``synthesis time'' indicates the time taken by PLA to tune the parameters. ``Total CPU time'' indicates the total time including model building and parsing time. 
The ``parameter space coverage'' is related to the approximation factor taken for PLA; it indicates the percentage of the entire parameter space covered by PLA for the synthesis. 
The column ``threshold found'' indicates the safe interval found by PLA for the given constraint.
For Bayesserver and SamIam, the columns ``threshold reported'' indicate the interval reported for the same pBN and constraint. 
The table also includes the ``tuning time'' and the ``total time'' for Bayesserver, which respectively are the time for the parameter tuning process and the total time --including the time for building and parsing the model. The synthesis time for Bayesserver is often less than Storm. This is because Bayesserver uses the existing closed-form formula to find the parameter instantiation for this specific subclass ($p_1c_1d_1$), while Storm does not exploit closed-form formulas and model-checks the entire parameter space. It determines the accepting and rejecting areas and is in principle applicable to arbitrary subclass $p_*c_*d_*$, see Tab. \ref{tab:pla-multiway}.  For two of the benchmarks Storm encountered time-out and memout while constructing the MC. \\
\indent Tab. \ref{tab:pla-multiway} (page \pageref{tab:pla-multiway}) indicates the PLA experiments performed on the \texttt{alarm} BN benchmark. 
The column ``$\#$parameterised rows'' indicates the number of CPT rows parameterized in the pBN, followed by the pBN class the benchmark belongs to. 
The subclass specifies the number of parameters, as well as the number of parametrized CPTs and the number of CPT rows the parameters are repeated at. The latter is related to the non-distinct parameters. 
The third column indicates the refinement factor taken for PLA. 
The fourth column indicates the number of regions synthesized by PLA; followed by the percentage of satisfied and unknown regions. 
The last four columns indicate the time taken by PLA for synthesis, the time taken by Storm for model building, the total process time, and the memory footprint. MO and TO respectively abbreviates ``memory-out'' and ``time-out'' of $15$ minutes in tables  \ref{tab:one-way}  and \ref{tab:pla-multiway} .

\begin{table}[htb]
    \centering
        \caption{Solution function calculation for the subclasses $p_1c_1d_1$, $p_2,c_{\leq2}d_1$, and $pBN$ - Storm vs Bayesserver}
        \label{sensitivity-table}
    \begin{tabular}{|c|c|c|c|c|}
       \toprule
         & \# parameters & Storm (regular) & Storm (bi-simulation) & Bayesserver \\
        \midrule
        cancer & 1 (one way) & 0.000s & \textbf{0.000s} & 0.011s \\
         & 2 (two way) & 0.000s & \textbf{0.000s} & 0.010s \\
         & 10 & 0.000s & \textbf{0.000s} & not supported \\ \hline
        earthquake & 1 (one way) & 0.000s & \textbf{0.000s} & 0.011s \\
         & 2 (two way) & 0.000s & \textbf{0.000s} & 0.011s \\
         & 10  & 0.000s & \textbf{0.000s} & not supported \\ \hline
        asia & 1 (one way) & \textbf{0.001s} & 0.003s & 0.0164s \\
         & 2 (two way) & 0.000s & \textbf{0.000s} & 0.0168s \\
         & 14  & 0.009s & \textbf{0.000s} & not supported \\ \hline
        sachs & 1 (one way) & 0.007s & \textbf{0.002s} & 0.0100s \\
         & 2 (two way) & 0.000s & \textbf{0.000s} & 0.0103s \\
         & 89 & 1.878s & \textbf{1.582s} & not supported \\ \hline
        alarm & 1 (one way) & 0.011s & \textbf{0.002s} & 0.018s \\
         & 2 (two way) & 0.025s & \textbf{0.004s} & 0.027s \\
         & 85 & 872s & \textbf{201s} & not supported \\ \hline
        insurance & 1 (one way) & 0.199s & 0.038s & \textbf{0.011s} \\
         & 2 (two way) & 0.178 & 0.036s & \textbf{0.034s} \\
         & 140 & 259s & \textbf{83s} & not supported\\ \hline
        win95pts & 1 (one way) & 0.002s & 0.\textbf{000s} & 0.0186s \\
         & 2 (two way) & 0.002s & \textbf{0.000s} & 0.0148s \\ 
         & 200 & 548s & \textbf{437s} & not supported \\ \hline
        hepar2 & 1 (one way) & 0.148s & 0.033s & \textbf{0.020s} \\
         & 2 (two way) & 0.130s & 0.026s & \textbf{0.020s} \\ 
         & 135 & 193s & \textbf{172s} & not supported \\ \hline
        hailfinder & 1 (one way) & 0.077s & \textbf{0.000s} & 0.009s \\
         & 2 (two way) & 0.080s & \textbf{0.000s} & 0.009s \\ 
         & 380 & 38s & \textbf{33s} & not supported \\ \hline
        water & 1 (one way) & 0.081s & \textbf{0.017s} & 0.058s \\
         & 2 (two way) & 0.089s & \textbf{0.012s} & 0.107s \\
         & 255  & \textbf{246s} & 247s & not supported \\
        \bottomrule
    \end{tabular}
\end{table}
\begin{landscape}
\begin{table}
\caption{Parameter tuning results for $p_1c_1r_1$ class on pBN benchmarks}
\label{tab:one-way}

\begin{adjustbox}{width=\columnwidth,center}
\begin{tabular}{|c|cccc|c|ccc|}
\hline
\multicolumn{9}{|c|}{Parameter Tuning (One Way)} \\
\hline
\multirow{2}{*}{ } & \multicolumn{4}{c|}{Storm} & SamIam & \multicolumn{3}{c|}{Bayesserver} \\
\hline
& synthesis time & total cpu time & parameter space coverage & threshold found & threshold reported & tuning time & total time & threshold reported \\
\hline
cancer & 0.001s & 0.028s & 99.9999\% & >=0.436054 & >=0.436054 & 0.0030s & 2.895s & >= 0.436054421768792 \\
earthquake & 0.001s & 0.029s & 99.99999\% & >= 0.268211 & >= 0.268211 & 0.0024s & 2.834s & >= 0.268211006844541 \\
asia & 0.001s & 0.029s & 99.99999\% & >= 0.294952 & >= 0.294952 & 0.0030s & 2.963s & >= 0.294952147495241 \\
sachs & 0.002s & 0.035s & 99.99999\% & >= 0.97554 & >= 0.97554 & 0.0028s & 2.920s & >= 0.975539734597883 \\
alarm & 0.016s & 0.087s & 99.9999\% & >= 0.362012 & >= 0.362012 & 0.0024s & 2.980s & >= 0.362012134508854 \\
barley & TO & TO & 90\% & - & >= 0.956102 & 0.0033s & 5.617s & >= 0.95610124817875 \\
insurance & 0.838s & 1.788s & 99.99999\% & >= 0.927801 & >= 0.927801 & 0.0033s & 3.012s & >= 0.927800679454384 \\
win95pts & 0.002s & 0.302s & 99.9999999\% & <= 0.274627 & <= 0.274627 & 0.0111s & 2.964s & <= 0.274626591335031 \\
hepar2-1 & 0.046s & 0.647s & 99.999999\% & <= 0.079602 & <= 0.079602 & 0.0027s & 3.056s & <= 0.0796019827080942 \\
hepar2-2 & 0.141s & 0.737s & 99.9\% & infeasible & infeasible & 0.0037s & 3.659s & infeasible \\
hailfinder-1 & 0.020s & 0.734s & 99.9999\% & >= 0.832838 & >= 0.832838 & 0.0099s & 3.456s & >= 0.8328381741961298 \\
hailfinder-2 & 0.021s & 0.796s & 99.9\% & infeasible & infeasible & 0.00348s & 3.668s & infeasible \\
water & 0.002s & 1.296s & 99.99999\% & <= 0.1 & <= 0.1 & 0.0035s & 3.492s & <= 0.09999999999999998 \\
andes & 5.456s & 346.056s & 99.999999\% & >= 0.8 & >= 0.8 & 0.0034s & 3.176s & >= 0.7999999999660568 \\
pathfinder & MO & MO & 90\% & - & <= 0.07 & 0.0029s & 4.260s & <= 0.06999999999999997 \\
\hline
\end{tabular}
\end{adjustbox}
\end{table}

\end{landscape}

\begin{landscape}
\begin{table}
\caption{Parameter tuning results for $p_*c_*r_*$ class on alarm network}
\label{tab:pla-multiway}
\begin{adjustbox}{width=\columnwidth,center}
\begin{tabular}{|c|c|c|c|c|c|c|c|c|c|}
\hline
\multicolumn{10}{|c|}{Parameter Tuning (Multiple-way) on Alarm network} \\
\hline
\#parameterised rows & pBN class & parameter space coverage & \#regions & satisfied & unknown & synthesis time & model building time & total time & memory \\
\hline
2 & p2c1r1 & 99.9\% & 6982 & 69.46\% & 0.10\% & 0.260s & 0.049s & 0.467s & 41MB \\
 2 & p1c1r2 & 99.9\% & 11 & 68.36\% & 0.10\% & 0.011s & 0.051s & 0.129s & 40MB \\
 2 & p2c2r1 & 99.9\% & 1 & infeasible & 0.00\% & 0.013s & 0.049s & 0.138s & 41MB \\
 3 & p2c2r2 & 99.9\% & 6982 & 69.47\% & 0.10\% & 0.293s & 0.047s & 0.499s & 41MB \\
 4 & p4c1r1 & 90\% & 73141 & 48.37\% & 10.00\% & 2.337s & 0.051s & 2.524s & 96MB \\
 4 & p4c1r1 & 99\% & - & - & - & - & TO & TO & -\\
4 & p3c1r2 & 90\% & 2640 & 48.24\% & 9.99\% & 0.158s & 0.052s & 0.279s & 42MB \\
 2 & p3c1r2 & 99\% & 232961 & 53.60\% & 1.00\% & 10.118s & 0.053s & 10.456s & 192MB \\
 4 & p1c1r4 & 99.9999\% & 22 & 53.12\% & 9.54E-07 & 0.023s & 0.051s & 0.144s & 41MB \\
 4 & p4c4r1 & 90\% & 6931 & 88.50\% & 9.99\% & 0.277s & 0.054s & 0.380s & 41MB \\
4 & p4c4r1 & 0.99\% & 10950661 & 94.55\% & 1\% & 294.248s & 0.055s & 308.285s & 8093MB \\
 4 & p1c4r4 & 99.9\% & 38 & 75.65\% & 5.96E-08 & 0.042s & 0.055s & 0.164s & 40MB \\
 6 & p6c1r1 & 80\% & 2928556 & 51.11\% & 20.00\% & 30.563s & 0.057s & 35.041 & 3291MB \\
6 & p4c1r3 & 90\% & 90316 & 56.60\% & 10.00\% & 1.967s & 0.052s & 2.146s & 108MB \\
 6 & p4c1r3 & 99\% & - & - & - & - & TO & TO & -\\
6 & p1c1r6 & 0.000001 & 22 & 58.62\% & 9.54E-07 & 0.022s & 0.052s & 0.148s & 41MB \\
8 & p8c1r1 & 60\% & 5397841 & 39.70\% & 40.00\% & 73.508s & 0.057s & 76.478s & 7308MB \\
8 & p5c1r4 & 90\% & 86776 & 57.14\% & 10.00\% & 2.067s & 0.054s & 2.249s & 106MB \\
8 & p1c1r8 & 99.9999\% & 21 & 58.88\% & 9.54E-07 & 0.024s & 0.054s & 0.151s & 41MB \\
8 & p8c4r1 & 60\% & 5906311 & 5.89\% & 40.00\% & 106.676s & 0.054s & 119.649s & 8084MB \\
 8 & p8c4r1 & 70\% & - & - & - & - & TO & TO & -\\
8 & p5c4r4 & 90\% & 2474421 & 19.96\% & 10.00\% & 81.685s & 0.053s & 85.196s & 2525MB \\
8 & p1c4r8 & 99.9999\% & 21 & 38.52\% & 9.54E-07 & 0.040s & 0.055s & 0.171s & 40MB \\
10 & p10c1r1 & 30\% & 205624 & 20.02\% & 79.98\% & 2.626s & 0.050s & 3.119s & 378MB \\
10 & p10c1r1 & 40\% & - & - & - & - & TO & TO & -\\
12 & p12c1r1 & 10\% & 3902536 & 10.01\% & 89.99\% & 61.229s & 0.051s & 73.949s & 7459MB \\
\hline
\end{tabular}
\end{adjustbox}
\end{table}

\end{landscape}

\end{document}